\documentclass[10pt,cspaper,compsoc]{IEEEtran}
\pdfoutput=1 
\usepackage{color}
\usepackage{graphicx}
\usepackage{cite}
\usepackage[cmex10]{amsmath}
\usepackage{amssymb}
\usepackage{subfigure}
\usepackage{rotating}
\usepackage{multirow}
\usepackage{makecell}
\usepackage{caption}
\usepackage{balance}
\usepackage{mathrsfs}
\usepackage{algorithm}
\usepackage{booktabs}
\usepackage{multicol}
\usepackage{float}
\usepackage{array}
\usepackage{colortbl}
\usepackage{siunitx}
\usepackage[keeplastbox]{flushend}
\usepackage{threeparttable}  % add footnotes under ta
\newcounter{magicrownumbers}  % number in table

\usepackage{ragged2e}

\def\ie{\emph{i.e.}}

\begin{document}

%\title{DiRS: On Creating Benchmark Datasets for  Remote Sensing Image Interpretation~{\textbf{\Large $^1$}}}
\title{Aerial Scene Parsing: From Tile-level Scene Classification to Pixel-wise Semantic Labeling*}

\author{Yang Long, Gui-Song~Xia, Liangpei Zhang, Gong Cheng, Deren Li

\IEEEcompsocitemizethanks{
%   \IEEEcompsocthanksitem{
% %\thanks{
% The study of this paper is funded by the National Natural Science Foundation of China (NSFC) under grant contracts No.61922065, No.61771350 and No.41820104006 and 61871299. It is also partially funded by the German Federal Ministry of Education and Research (BMBF) in the framework of the international future AI lab ``AI4EO -- Artificial Intelligence for Earth Observation: Reasoning, Uncertainties, Ethics and Beyond". } % <-this % stops a space
\IEEEcompsocthanksitem{
%\thanks{
Y. Long, L. Zhang, D. Li are with the State Key Lab. LIESMARS, Wuhan University, Wuhan, China. 
e-mail: \{{\em longyang, zlp62, drli}\}@whu.edu.cn}% <-this % stops a space
\IEEEcompsocthanksitem{
%\thanks{
G.-S. Xia is with the School of Computer Science and also the State Key Lab. LIESMARS, Wuhan University, Wuhan, China. 
e-mail: guisong.xia@whu.edu.cn}% <-this % stops a space
\IEEEcompsocthanksitem{
%\thanks{
Gong Cheng is with the School of Automation, Northwestern
Polytechnical University, Xi’an, China.
e-mail: gcheng@nwpu.edu.cn}% <-this % stops a space
\IEEEcompsocthanksitem{
%\thanks{
The corresponding author is Gui-Song Xia (guisong.xia@whu.edu.cn).}
\IEEEcompsocthanksitem[{\textbf *}]{
%\thanks{
A website is available at: {\textcolor[RGB]{236,0,140}{https://captain-whu.github.io/ASP/}}}
}
}

\IEEEtitleabstractindextext{
\begin{abstract}
\justifying
Given an aerial image, aerial scene parsing (ASP) targets to interpret the semantic structure of the image content, {\em e.g.}, by assigning a semantic label to every pixel of the image. With the popularization of data-driven methods, the past decades have witnessed promising progress on ASP by approaching the problem with the schemes of {\em tile-level scene classification} or {\em segmentation-based image analysis}, when using high-resolution aerial images. 
However, the former scheme often produces results with tile-wise boundaries, while the latter one needs to handle the complex modeling process from pixels to semantics, which often requires
large-scale and well-annotated image samples with pixel-wise semantic labels.
In this paper, we address these issues in aerial scene parsing, with perspectives from tile-level scene classification to pixel-wise semantic labeling. 
% To overcome this problem, we propose to perform aerial scene parsing by unifying the tile-level scene classification and object-based image analysis to achieve pixel-level semantic labeling.
Specifically, we first revisit aerial image interpretation by a literature review. We then present a large-scale scene classification dataset that contains one million aerial images termed Million-AID. With the presented dataset, we also report benchmarking experiments using classical convolutional neural networks (CNNs). Finally, we perform ASP by unifying the tile-level scene classification and object-based image analysis to achieve pixel-wise semantic labeling. 
Intensive experiments show that Million-AID is a challenging yet useful dataset, which can serve as a benchmark for evaluating newly developed algorithms. When transferring knowledge from Million-AID,
fine-tuning CNN models pretrained on Million-AID perform consistently better than those pretrained ImageNet for aerial scene classification, demonstrating the strong generalization ability of the proposed dataset. Moreover, our designed hierarchical multi-task learning method achieves the state-of-the-art pixel-wise classification on the challenging GID, which is a profitable attempt to bridge the tile-level scene classification toward pixel-wise semantic labeling for aerial image interpretation. We hope that our work could serve as a baseline for aerial scene classification and inspire a rethinking of the scene parsing of high-resolution aerial images. 
\end{abstract}

\begin{IEEEkeywords}
Aerial image interpretation, Million-AID, scene classification, semantic segmentation, transfer learning
% \texttt{elsarticle.cls}\sep \LaTeX\sep Elsevier \sep template
% \MSC[2010] 00-01\sep  99-00
\end{IEEEkeywords}}

\maketitle
\IEEEdisplaynontitleabstractindextext

\section{Introduction} \label{Introduction}
Aerial image understanding is a task of primary importance for a wide range of applications such as precision agriculture~\cite{weiss2020remote}, urban planning~\cite{wellmann2020remote}, and environmental monitoring~\cite{yuan2020deep}. An essential way toward understanding an aerial image is to perform a full-scene semantic structure interpretation, also denoted as aerial scene parsing (ASP), which aims to label each pixel in the image with a semantic category to which it belongs. With more and more aerial images being available, ASP has been a momentous but active topic in the field of remote sensing~\cite{porway2010hierarchical,zheng2020parsing,zhou2021cegfnet,wang2021fully}. Besides, pixel-wise semantics acquired by ASP is usually demanded as the imperative prerequisite in practical applications like land use/land cover investigation~\cite{Tong2019GID,9027099,zhiyong2021land}. However, aerial images taken from the bird's view with large imaging angles are always characterized with large scale, which implies that conventional computational method with pixel-wise analysis is hard to fulfill a full aerial scene parsing. Moreover, the highly complex content and image structure further increase the difficulty of identifying the semantics of pixels in an aerial image.

\begin{figure*}
    \centering
    \includegraphics[width=.95\linewidth]
      {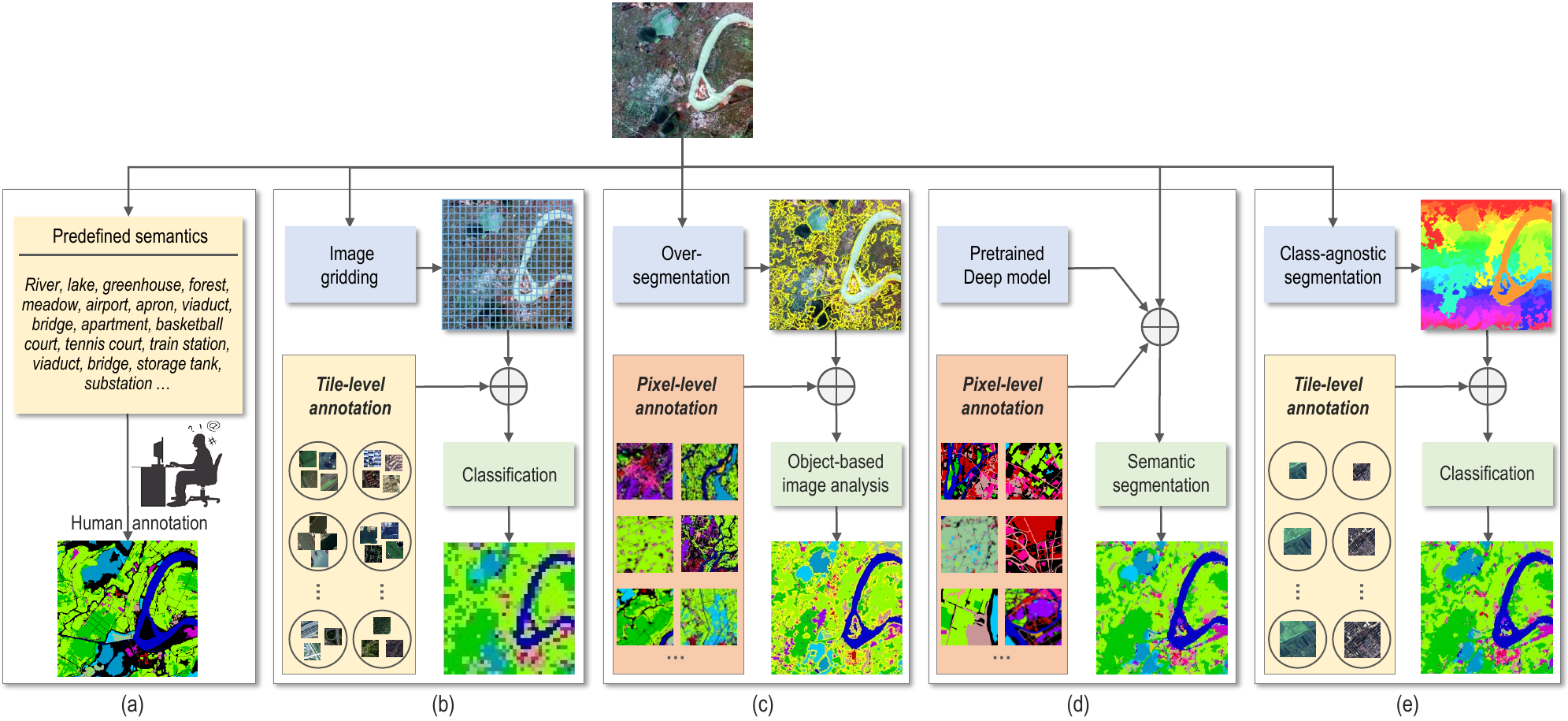}
    % \vspace{-1mm}
    \caption{Aerial scene parsing implemented by (a) human annotation, (b) tile-level scene classification, (c) object-based image analysis, (d) end-to-end semantic segmentation, and (e) our proposed method. Our basic idea is to utilize multi-scale contextual information and hierarchical convolutional features with tile-level scene classification, where image segmentation is incorporated to extract homogeneous region for producing pixel-wise semantic boundaries.}
    \label{fig:header}
    % \vspace{-4mm}
\end{figure*}

Faced with this situation, aerial scene parsing has been simplified as tile-level scene classification which integrates the complicated image features and content as a whole for semantic classification~\cite{UCM,sheng2012high,hu2015transfer,Yang2015Scene,pyramid2015chen,AID,RESISC45,toward2017nogueira,cheng2020deepsc,chen2021drsnet}. However, the tile-level scene classification focuses on summarizing the thematic content within a local area while accurate semantics of individual pixels cannot be identified. To remedy this defect, the scene image can be divided into different regions for semantic identification.
A feasible way is to split the scene image into regular grids, on which the patch-based classification can be performed for aerial scene parsing ~\cite{sharma2017patch,paoletti2018new,SHARMA2018346,liu2020local}. Nevertheless, it usually produces coarse maps with blurred object edges.
Moreover, object-based image analysis (OBIA) that extracts homogeneous entities by over-segmentation has been widely employed to achieve pixel-wise semantic labeling~\cite{blaschke2001s,blaschke2010object,blaschke2014geographic,hossain2019segmentation,martins2020exploring}. Even with great success, aerial scene parsing relies on OBIA suffers from parameter optimization~\cite{MING201528,MA201514}, feature selection~\cite{ma2017review}, and modeling the complex relationship among different objects when reasoning their semantic meaning~\cite{porway2010hierarchical}. 
Recently, semantic segmentation based on fully convolutional networks (FCNs)~\cite{long2015fully,UNet2015,segnet2017,chen2017deeplab} provides an end-to-end aerial scene parsing framework that has been intensively approached~\cite{deepreview,li2019deep,audebert2019deep,zheng2020parsing,zhou2021cegfnet,wang2021fully,yang2021hidden}. Still, the optimization of FCN methods requires a large amount of pixel-wise annotation which is extremely labor-intensive and time-consuming to produce. Figure~\ref{fig:header}(a)$\sim$(d) illustrate the conventional methods for aerial scene parsing by human annotation, tile-level classification, object-based analysis, and end-to-end semantic segmentation, respectively. From an overall perspective, the above methods for aerial scene parsing are typically performed in a separate way, of which advantages cannot be integrated. To change this situation, a great deal of effort must be paid by addressing the following critical issues: 

 \begin{itemize} 
    \item \emph{The divergence of aerial scene parsing prototypes in tile-level and pixel-wise classification.} Pixel-wise classification has been intensively approached to produce fine-grained labeling result~\cite{lu2007survey,CHEN20082731,tuia2011survey} while tile-level classification can only provides coarse semantic description of regions~\cite{UCM,Yang2015Scene,RESISC45,AID,cheng2020deepsc}. Currently, the continuous improvement of image resolution has brought us aerial images of large scale and rich detail. As a result, aerial scene parsing by pixel-wise classification becomes a challenging task because of the variant attributes of pixels and the high computational cost. From the perspective of image expression, the improvement of image resolution also greatly enhances the semantic homogeneity of pixels in local regions. Thus, the semantics of individual pixels in an aerial image is closely related to their contextual information rather than relying solely on their own. In this context, it is reasonable to consider the pixel as a unit centered by tile-level scene and bridge its gap to pixel-wise semantic labeling.
     
     \item \emph{The scarcity on exploring the transferability of semantic scene knowledge of aerial images.} Currently, the lack of large-scale benchmark datasets has become a bottleneck that hampers the development of data-driven methods for aerial image interpretation, particularly the deep learning-based ones. To alleviate this situation, the conventional way is to employ CNN models pretrained on natural image archives (\emph{e.g.}, ImageNet~\cite{Imagenet}) as feature extractors or to fine-tune them on target aerial images. However, there are great differences between natural and aerial images in spectral properties, image structure, and spatial arrangement. Moreover, the semantic categories that define the scene content also vary significantly between natural and aerial images. Thus, the learned features obtained through the above methods can be biased in characterizing aerial image content. With this in mind, it is of great significance to explore the transferability of data-driven models adapted with pure aerial scene images and free up their potential for aerial image interpretation. 
 \end{itemize}
 
 With these points in mind, this paper addresses aerial scene parsing that unifies tile-level scene classification and object-based image analysis to achieve pixel-wise semantic labeling as illustrated in Figure~\ref{fig:header}(e). For doing so, we first provide a review to depict aerial image interpretation. Then, we present a large-scale aerial scene classification dataset, \emph{i.e.}, Million-AID, on which the benchmarking experiments are performed to investigate how well the current CNNs perform for aerial scene classification. Finally, we conduct aerial scene parsing from tile-level scene classification to pixel-wise semantic labeling, where knowledge transferring from Million-AID is performed to improve the accuracy. To sum up, our main contributions are as follows: 
 
  \begin{itemize}
    \item We provide a comprehensive literature review on aerial image interpretation by revisiting its development outline, ranging from pixel-wise image classification, segmentation-based image analysis, and tile-level image classification that are connected to the improvement of spatial resolution of aerial images.
  
    \item We released a new large-scale dataset, \ie, Million-AID, for aerial scene recognition and benchmark a number of classical CNN models on multi-class and multi-label aerial scene classification. To the best of our knowledge, Million-AID is one of the largest benchmark datasets in the remote sensing community. The experimental results also demonstrate that Million-AID is challenging but useful, and can serve as a large-scale benchmark for developing aerial scene classification algorithms. 
    
    \item We conducted extensive experiments to verify the tremendous potential of transferring scene knowledge from tile-level annotated Million-AID to pixel-wise semantic labeling for aerial scene parsing. {\em Experiments show that fine-tuning CNN models pretrained on Million-AID perform consistently better than those pretrained on ImageNet for aerial scene classification.} Our proposed hierarchical multi-task learning method exploiting Million-AID and GID achieves the state-of-the-art on pixel-wise aerial image classification.
  \end{itemize}
  
  The remainder of this paper is organized as follows. Section~\ref{Review} presents the review of aerial image interpretation. Section~\ref{MAIDIntroduction} introduces the proposed large-scale scene classification dataset, \emph{i.e.}, Million-AID. Section~\ref{MultiClassSC} presents the comprehensive benchmarking experiments on Million-AID, including multi-class and multi-label aerial scene classification. Section~\ref{KnowledgeTransfer} presents experiments tile-level scene classification and pixel-wise semantic labeling with knowledge transfer from Million-AID. Finally, in Section~\ref{Conclusions}, we draw conclusions regarding this work. 

\section{Revisiting Aerial Image Interpretation} \label{Review}

 \begin{figure*}
      \centering
      \includegraphics[width=0.99\linewidth]
      {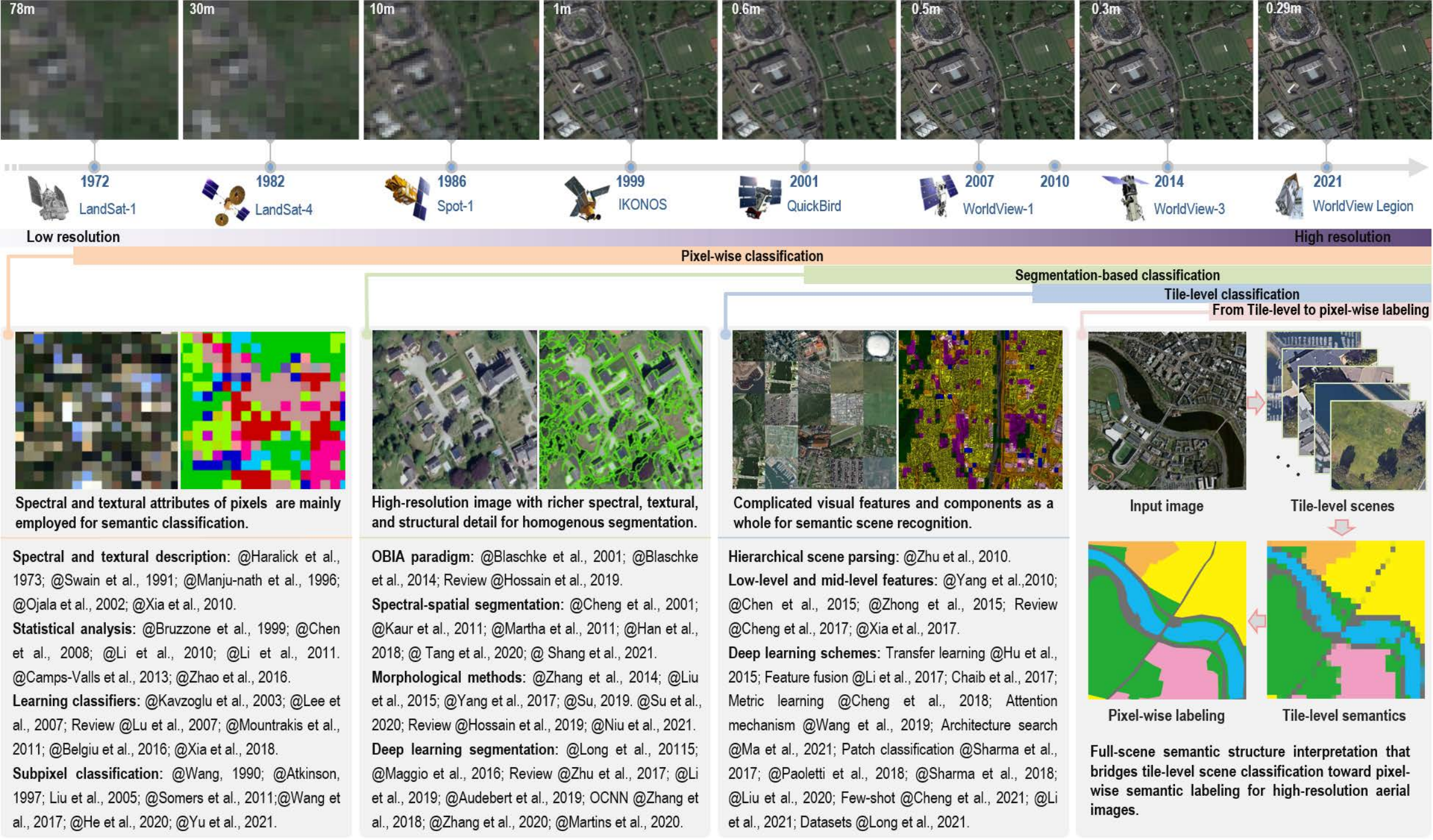}
      \caption{The literature review of aerial image interpretation. The interpretation prototypes develop with the spatial resolution improvement of aerial images and have experienced a long course of development ranging from pixel-wise classification, segmentation-based classification, to tile-level classification. This figure non-exhaustively presents some representative works concerning aerial image recognition. In this work, we will make our efforts to perform semantic classification that bridges tile-level scene classification toward pixel-wise semantic labeling for high-resolution aerial scene parsing. Zoom for detail.}
      \label{figure:RoadMap}
      \vspace{-4mm}
  \end{figure*}

With the progress of sensor technology, the spatial resolution of aerial images has experienced a continuous improvement~\cite{Toth201622,PU2012516}. Figure~\ref{figure:RoadMap} presents the milestones of earth observation satellites at different times. Accordingly, the improvement of aerial image resolution has greatly promoted the development of aerial image interpretation. 
% Accordingly, the interpretation of aerial images has experienced a long course of development ranging from per-pixel classification, object-based analysis, to scene-level understanding. 
In this section, we focus on a review by revisiting the interpretation tasks of aerial images and the outline is presented in Figure~\ref{figure:RoadMap}.

\subsection{Pixel-wise aerial image classification} 
In the early 1970s, aerial images are characterized with a low spatial resolution (\emph{e.g.}, LandSat-1 and MODIS images) where each pixel describes an area of thousands of square meters of the Earth's surface. And the sizes of ground features or objects are usually smaller than the ground sampling distance of image pixels. Thus, each pixel is able to present a scene of a specific semantic category. Individual pixels are obviously distinct from each other owing to the difference in covered ground features. In this situation, semantic interpretation of aerial images mainly focuses on pixel-wise classification using spectral signatures~\cite{swain1991color,ojala2002multiresolution} and coarse textural features~\cite{haralick1973textural,manjunath1996texture,ojala2000gray,xia2010shape}. To this end, sampling analysis is naturally employed~\cite{VANGENDEREN19783,curran1986sample,KHATAMI2017156} to construct desirable classification schemes. Typically, training samples that are representative to reflect the distribution and variation of the diverse semantic content of interest are selected to extract category information. Thus, classification methods based on statistical analysis are widely employed by estimating the probability of a pixel belonging to each of the possible classes~\cite{bruzzone1999neural,CHEN20082731,li2010semisupervised,camps2013advances,li2011hyperspectral,zhao2016high}. 

In order to obtain reliable classification results, a number of classifiers were also developed for the pixel-wise aerial image parsing, such as maximum likelihood methods~\cite{settle1987fast,ediri1997mlc,peng2019mlc}, minimum distance to means algorithms~\cite{HODGSON1988117,espinola2014contextual}, K-nearest neighbors classifiers~\cite{ma2010local,tu2018knn}, and tree-based techniques~\cite{friedl1997decision,otukei2010land}.
However, statistical classification methods usually show insufficient ability in discriminating pixel units due to the variation of spectral and spatial characteristics influenced by imaging conditions and ground feature attributes. Faced with this situation, more sophisticated classifiers such as random forest~\cite{BELGIU201624,xia2018random,izquierdo2020evaluation,zafari2020multiscale}, sparse representation~\cite{lee2007efficient,liu2013spatial,feng2017superpixel,fan2017superpixel,peng2019mlc,li2020adaptive}, artificial neural network~\cite{kavzoglu2003use,SHAO201278}, and kernel-based methods~\cite{liu2013spatial,gu2017multiple,li2020adaptive,zafari2020multiscale} represented by support vector machine~\cite{mountrakis2011support,maulik2017remote,okwuashi2020deep} were intensively explored by actively embracing the machine learning techniques. These methods have made dramatic progress in pixel-wise aerial image classification owing to their strong ability to discriminate the complex spectral and spatial characteristics of ground features.  

Pixel-wise classification approaches assume that each pixel only belongs to a single semantic category and different categories are mutually exclusive. However, such an assumption can be inconsistent with reality due to the limitation of spatial resolution of aerial images~\cite{lu2007survey,li2014review}. Specifically, more than one object or ground feature belonging to different semantic categories can be contained within a pixel as their scales are smaller than the spatial resolution of aerial images. As a result, the existence of mixed pixels comes to be a nonnegligible problem in the medium and coarse spatial resolution aerial images. This could lead to an appreciable interpretation result when employing the pixel-wise classification strategy. To overcome this problem, sub-pixel classification is considered as an alternative for more accurate aerial image interpretation~\cite{liu2005comparison,bovolo2010novel,wang2020sub,he2021supixel}. 

A number of approaches have been derived to address the sub-pixel aerial image classification, including soft or fuzzy theory~\cite{wang1990fuzzy,shackelford2003hierarchical,kothari2020improved}, neural networks~\cite{mertens2004sub,li2014spatial,he2020subpixel}, regression modeling and analysis~\cite{fernandes2004approaches,gessner2013estimating,cooper2020disentangling}, and spectral mixture analysis~\cite{wu2003estimating,somers2011endmember,yu2021novel,xu2021unmixing}. Among these methods, the fuzzy technique and spectral mixture analysis are most commonly employed to overcome the mixed pixel problem. Particularly, fuzzy representation is developed to estimate multiple and partial memberships of all candidate categories within a pixel, where the corresponding areal proportion of each category can be acquired. The spectral mixture analysis assumes the value of a pixel is a linearly or non-linearly combination of a set of specific endmember spectra~\cite{somers2011endmember,yu2021novel}. Thus, the selection of endmembers becomes the key point for designing an effective classifier~\cite{bateson1996method,small2004landsat,EndNet2019,hong2021endmember}. Even with a great improvement of classification accuracy, sub-pixel class composition estimated by fuzzy classification and spectral mixture analysis cannot provide the spatial distribution of land cover classes within pixels. To address this issue, the sub-pixel mapping approaches are developed~\cite{atkinson1997mapping,wang2017effect,wang2020sub,he2020subpixel}. In this scheme, each pixel is divided into sub-pixels which are predicted to get single semantic labels. Limited by the spatial resolution, aerial images interpreted by pixel-wise classification still face challenges in acquiring satisfactory results due to the mixture and complexity of image content within single pixels. 

\subsection{Segmentation-based aerial image analysis} 
With the development of sensor technology, the spatial resolution of aerial images is gradually improved to be much smaller than the scales of ground features. Thus, the detail of spectrum, texture, and geometric structure in the image becomes prominent. Under the circumstances, single pixels are no longer isolated units since the ground features and objects could be composed of a certain number of pixels knitted into an image full of spatial patterns~\cite{hay2008geographic}. And the improved image quality also significantly increases the within-class variability, which decreases the potential accuracy of pixel-based approach to classification~\cite{blaschke2014geographic}. As a result, the traditional interpretation system established with pixel-wise statistics and analysis for low-resolution aerial images, to some extent, is beginning to show cracks in classifying aerial images for required accuracy and generalization ability~\cite{blaschke2010object}. 

Faced with this situation, researchers turn their attention to the new paradigm of object-based image analysis (OBIA) or geographic-object-based image analysis (GEOBIA)~\cite{blaschke2001s,hay2008geographic,blaschke2010object,blaschke2014geographic,ma2017review,hossain2019segmentation,kotaridis2021remote}, where geographical or image objects are considered as the basic units instead of individual pixels for image classification. The objects are considered to be homogeneous entities, located within an image and perceptually generated from pixel-groups, where each pixel-group is composed of similar digital values and possesses an intrinsic size, shape, and geographic relationship with the real-world scene component it models~\cite{hay2001multiscale}. In general, the OBIA generates objects by image segmentation and then performs image classification on objects. Thus, image segmentation serves as the initial and critical part to produce the fundamental elements of OBIA~\cite{martins2020exploring,kotaridis2021remote}.

In high-resolution remote sensing images, ground objects are presented with much richer spectral, textural, structural, and contextual detail that reveals the pattern characteristics~\cite{blaschke2014geographic}. It enables the objects of interest to be extracted by spectrally-based and spatially-based segmentation approaches, among which mathematical morphology analysis plays a significant role~\cite{cheng2001color,hossain2019segmentation}. Hence, the thresholding~\cite{5783913,YANG2017137,tang2020integrating} and feature space clustering~\cite{amitrano2018feature} methods are typically employed to generate objects by spectral analysis based on the fact that homogeneous objects share similar spectral characteristics. 
% For spatially-based segmentation, edge detection~\cite{kaur2011mathematical,YANG2017137,han2018edge,shang2021sar}, region generation (\emph{e.g}, region growing~\cite{5783913,LIU2015145}, merging, and splitting~\cite{YANG2017137,su2019scale,su2020machine}), hybrid segmentation~\cite{ZHANG201419,YANG2017137,niu2021hybrid} techniques are conducted according to the discontinuity and similarity of object areas.  
For spatially-based segmentation, edge detection~\cite{kaur2011mathematical,YANG2017137,han2018edge,shang2021sar} and region generation (\emph{e.g}, region growing~\cite{5783913,LIU2015145}, merging, and splitting~\cite{YANG2017137,su2019scale,su2020machine}) techniques are conducted according to the discontinuity and similarity of object areas, respectively. However, edge-based methods are precise in boundary detection while facing problems in generating closed segments. By contrast, region-based methods have the advantage in generating closed regions while resulting in imprecise segment boundaries. As compensation, hybrid segmentation~\cite{ZHANG201419,YANG2017137,niu2021hybrid} that consider both the boundary and spatial information between adjacent regions show significant advantages in object segmentation. 

However, objects acquired by segmentation can only provide homogeneous regions while lacking a semantic description. Thus, the object features are then extracted and embedded into a classifier to determine the semantic categories. In doing so, both the feature extraction and classifier design play crucial roles in classification performance~\cite{ghamisi2017advanced,kumar2020feature}. Owing to the overwhelming advantages in visual feature extraction and classification, CNN frameworks have recently been integrated into OBIA and triggered the new trend of object-based CNN (OCNN) for aerial image interpretation~\cite{zhang2018object,zhang2020multi,martins2020exploring}. With the availability of high-resolution aerial images, object-based approaches become dominant in the task of aerial image interpretation over the past two decades. Even with a significant performance advantage compared with pixel-wise classification methods, the object-based classification methods face challenges in parameter setting and optimization (\emph{e.g.}, segmentation scale)~\cite{MING201528,MA201514}, which affect the segmentation quality as well as the final classification accuracy. In addition, the segmentation and classification of objects fall into a complicated and multi-step implementation pipeline, which inevitably limits the efficiency and increases the difficulty of model deployment. 

To overcome the above deficiencies, the solution that simultaneously produces the segmented homogeneous areas and corresponding semantic categories becomes an imperious demand. In recent years, aerial image classification have been greatly facilitated by deep convolutional neural networks (DCNNs)~\cite{maggiori2016convolutional,li2019deep,audebert2019deep,jia2021survey}, among which the fully convolutional network (FCN)~\cite{long2015fully} and the improved 
architectures~\cite{UNet2015,segnet2017,chen2017deeplab} provide an end-to-end segmentation and classification pipeline. In contrast with conventional methods, the advantage of DCNN lies in its capacity to extract shallow visual and deep semantic features by the elaborately designed hierarchical framework~\cite{deepreview}. However, the down-sampled features in deep layers will lead to the resolution degradation for the final classification results, in which the uncertainty of boundaries and detail of different classes is a serious issue. To deal with these issues, the multi-scale features ~\cite{zhang2019multi,yang2019multi,peng2019densely,sun2020deep} and contextual information~\cite{zhang2018diverse,zhang2020multi} are typically considered to enhance the feature representation ability. The atrous and paralleled dilation convolution~\cite{chen2017deeplab} is utilized to preserve feature resolution~\cite{niu2018deeplab}. 
% And object-oriented CNNs are also incorporated into object-based image analysis for better discrimination of object boundaries~\cite{zhang2018object,zhang2020multi}. 
With the improvement of network architectures, the classification accuracy of aerial images has witnessed a continuous improvement. 

Apart from the evolution of networks, significant efforts have been paid to improve the performance of classification models. Typically, the spatial and spectral attributes of aerial images have been intensively addressed in pixel-wise classification ~\cite{he2017recent,ghamisi2018new,gao2019spectral}. Particularly, the integration of spatial and spectral information is commonly employed to address the challenge of large spatial variability of spectral signatures~\cite{mou2019learning,feng2019cnn,li2020deep,imani2020overview}. 
% To take full advantage of different networks, the ensemble of multiple networks is also explored for feature complementary and has achieved encouraging performance~\cite{chen2019deep,he2020transferring}. 
Regarding the limitation of training samples and generalization ability, transfer learning has been intensively explored to address the limitation of training samples for CNN frameworks and reported promising classification results~\cite{Tong2019GID,wurm2019semantic,chen2019deep,he2020transferring}. The readers may go to one of the review papers for a more comprehensive perspective of semantic segmentation using deep learning techniques~\cite{deepreview,he2017recent,ghamisi2018new,ma2019deep,imani2020overview,lateef2019survey,minaee2021image}.
However, owing to the lack of large-scale datasets, many interpretation algorithms are locally-oriented, typically manifested in the validation of one or several images within local areas which would affect the generalization ability. And the CNN-based methods also suffer from computational burden when classifying aerial images of large size and huge volume due to the improvement of spectral and spatial resolution. 

 \subsection{Tile-level aerial image classification}
Even with the impressive success achieved by object-based analysis, individual objects carry information independent to their neighbors and thus neglect the thematic meaning in their contextual environment, which could lead to inaccurate classification results. To alleviate this problem, the hierarchical and contextual model is developed by organizing individual objects into hierarchical groups for aerial image parsing~\cite{porway2010hierarchical}. However, the implementation of object detection and hierarchical contextual representation is complicated. Thus, the classification of the tile-level scene, which is able to incorporate visual features, content components, and spatial arrangements as a whole, becomes an effective way for aerial image interpretation. In the last decade, a handful of visual descriptors have been employed for aerial scene classification~\cite{UCM,xia2010structural,pyramid2015chen,topic2015zhong,RESISC45,AID,cheng2020deepsc}. We refer interested readers to~\cite{UCM,RESISC45,AID} for a survey of the low-level and middle-level visual features employed for aerial scene classification. 

Owing to the increasing accessibility of aerial images, the data-driven approaches particularly CNN-based ones have shown a great advantage over the handcrafted-feature-based approaches for aerial scene classification. In the beginning, pretrained CNNs are usually employed as feature extractors owing to their simplicity and efficiency~\cite{hu2015transfer,mlf2017li,mls2018he}. However, the representation of aerial scenes is a challenging task owing to the complexity of scene components and scale variation. To improve the feature discrimination ability, multi-scale images or features are extracted and fused to generate robust global representation for scene classification~\cite{mlf2017li,chaib2017deep,mls2018he,caps2020mulconv,bi2021multi}. In fact, the semantic category of an aerial scene usually depends on the spatial arrangement and class-specific objects in the image. Thus, deep local structures related to scene category are addressed to improve the classification performance~\cite{rearrange2019yuan,local2021bi,qi2022allgain}. Recently, CNNs based on attention mechanism have been addressed to highlight more local semantics and discard the noncritical information~\cite{rcurr2019atten,bi2020multiple,recurr2020atten,bi2021multi}. In general, learning powerful features for content representation is of great importance for aerial scene recognition. 

With the improvement of spatial resolution of aerial images, the within-class diversity and between-class similarity of semantic scenes have been greatly increased, which makes scene recognition a challenging task. To relieve this issue, deep metric learning algorithms are developed to learn discriminative category features~\cite{metric2018cheng,metric2018gong,metric2020kang}. Particularly, the complex relationship pervading aerial scenes is further explored in the embedding space by learning deep graph networks~\cite{LRAGE2018wang,graph2019khan,graph2021kang}. The main idea is to map the scene features closely to each other for the same categories while as farther apart as possible for different categories.
As conventional CNNs with a fixed architecture may show limitations in grasping the scene content of large diversity, automatically learning CNN architecture specified for aerial scenes has been intensively explored~\cite{ma2021scenenet,cnnsearch2021peng,cnnsearch2021broni} and achieved encouraging results. 

However, the methods based on deep learning require large-scale annotated samples for model adaption while most of them are trained and tested on relatively small-scale datasets. To overcome this problem, scene classification based on few-shot learning has recently attracted extensive attention~\cite{fewshot2021cheng,fewshot2021li,fewshot2021liu}. Moreover, annotated scene images from different domains are also employed to relieve the issue of data dependency by transfer learning~\cite{domain2017othman,domain2020lu,domain2020zhang,AMSCS2021ZH}. These approaches have reported exciting performance on aerial scene classification. However, the recent scene classification algorithms have intensively reported saturation results as shown in~\cite{cheng2020deepsc}. Faced with this situation, the potential of the data-driven methods for scene classification remains to be further explored and boosted by large-scale datasets. 

The aforementioned prototypes have achieved great success in aerial image interpretation. However, the conventional OBIA method usually results in a complicated modeling process for semantic reasoning while end-to-end semantic segmentation requires large-scale pixel-wise annotations for model adaption. The tile-level scene classification also shows a deficiency in identifying the semantics of individual pixels. As a compromise, patch-based classification has been intensively approached by shrinking the scale of an aerial scene~\cite{sharma2017patch,paoletti2018new,SHARMA2018346,liu2020local}. Nevertheless, it produces the classification map with blurred semantic boundaries. From a geographical perspective, the discrimination of a ground object relies heavily on the background environment. Thus, the semantics of individual pixels are closely related to their surrounding neighbors in high-resolution aerial images. In this situation, it is reasonable to employ the tile-level scene to incorporate the contextual information of its central pixels and then predict the semantic meaning. And the acquisition of tile-level scene labels is much easier than those of pixel-wise ones. With these in mind, we aim to unify the aforementioned prototypes and perform aerial scene parsing from tile-level scene classification to pixel-wise semantic labeling in this work. 

\section{An Introduction to Million-AID} \label{MAIDIntroduction}
% Due to the difficulty of aerial image collection and high cost of semantic annotation, there is a lack of large-scale aerial scene datasets which play a vital role in the development of aerial image interpretation algorithms. 
In this section, we detail the properties of Million-AID to be released for aerial scene classification. The readers may go to~\cite{Long2020DiRS} for the construction of Million-AID. 

  \begin{figure*}
      \centering
      \includegraphics[width=0.8\linewidth]
      {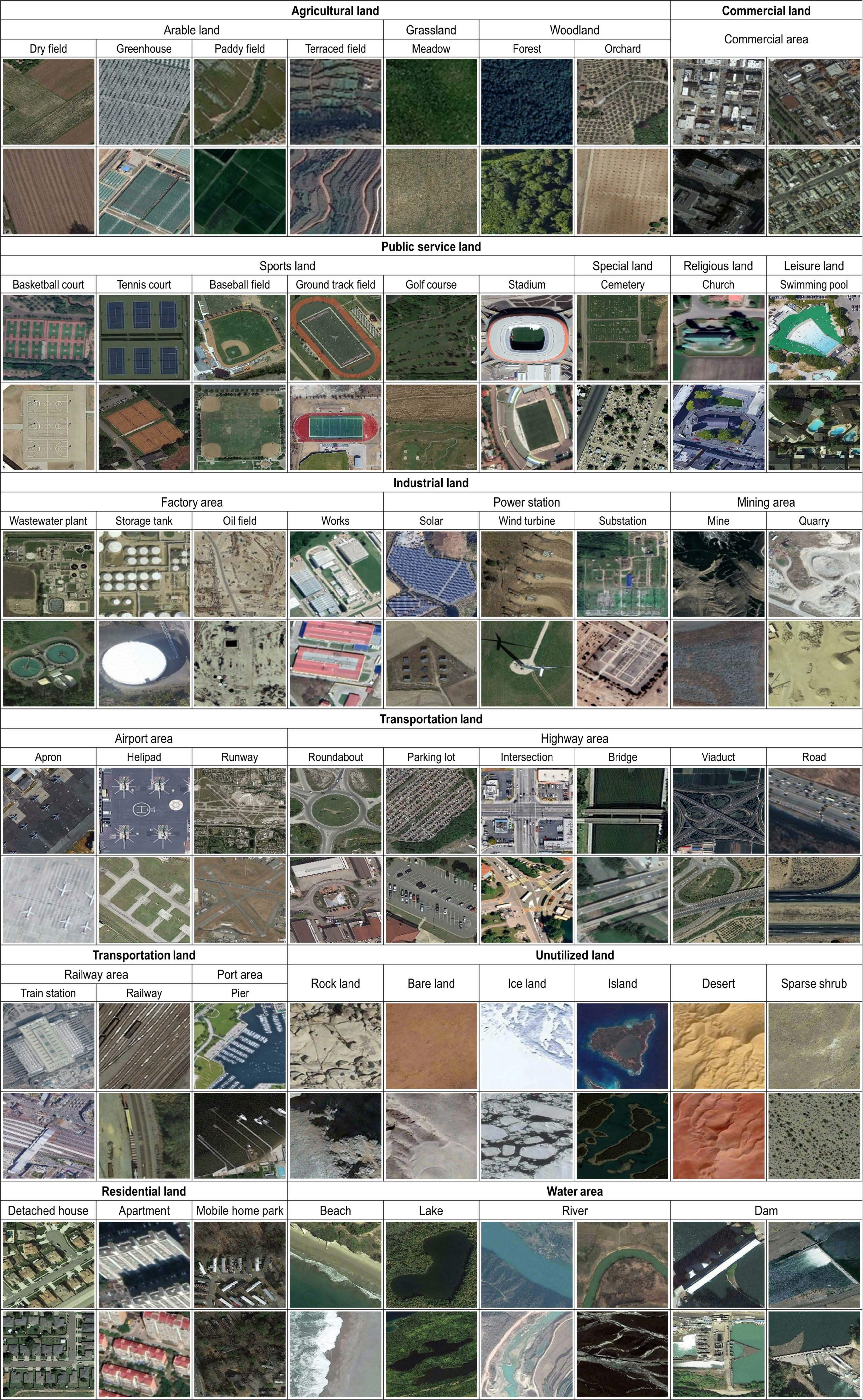}
      \caption{Scene samples of Million-AID: two or four examples of each semantic category are presented. All the semantic scenes are organized by the hierarchical system with three-level semantic labels, containing 51 fine-grained scene categories belonging to 8 major categories.}
      \label{figure:MAID_Samples}
      \vspace{-3mm}
  \end{figure*}
  
  \subsection{Scene categories} 
  The semantic scenes in Million-AID are hierarchically organized by referencing the land-use classification standards. There are 8 major classes of aerial scenes in the first level,~\emph{i.e.},~\emph{agriculture land},~\emph{commercial land},~\emph{public service land},~\emph{industrial land},~\emph{transportation land},~\emph{residential land},~\emph{water area}, and~\emph{unutilized land}, covering 28 sub-classes in the second level. And more specific scene categories are organized at the third level. In total, there are 51 fine-grained scene categories, including~\emph{dry field} (DF),~\emph{greenhouse},~\emph{paddy field} (PF),~\emph{terrace field} (TF),~\emph{meadow},~\emph{forest},~\emph{orchard},~\emph{commercial area} (CA),~\emph{storage tank} (ST),~\emph{wastewater plant} (WP),~\emph{works, oil field},~\emph{mine},~\emph{quarry},~\emph{solar power plant} (SPP),~\emph{wind turbine} (WT),~\emph{substation},~\emph{swimming pool} (SP),~\emph{church},~\emph{cemetery},~\emph{basketball court} (BC),~\emph{tennis court} (TC),~\emph{baseball field} (BF),~\emph{ground track field} (GTF),~\emph{golf course} (GC),~\emph{stadium},~\emph{detached house} (DH),~\emph{apartment},~\emph{mobile home park} (MHP),~\emph{apron},~\emph{helipad},~\emph{runway},~\emph{road},~\emph{viaduct},~\emph{bridge},~\emph{intersection},~\emph{parking lot},~\emph{roundabout},~\emph{pier},~\emph{railway},~\emph{train station} (TS), ~\emph{rock land},~\emph{bare land},~\emph{ice land},~\emph{island},~\emph{desert},~\emph{sparse shrub land} (SSL),~\emph{lake},~\emph{river},~\emph{beach}, and~\emph{dam}. All labels have been checked by the specialists in the field of aerial image interpretation. Several instances in each scene class are shown in Figure~\ref{figure:MAID_Samples}. Moreover, each scene image can be assigned with more than on category labels according to the hierarchical semantic nodes.
%   Moreover, as the scenes in Million-AID are organized hierarchically as mentioned before, each scene image can be assigned with more than on category labels according to the hierarchical semantic nodes.
% More details of the scene category organization are available in~\cite{Long2020DiRS}.
% Moreover, as the scenes in Million-AID are organized with a hierarchical network as mentioned before, each scene image that falls into a leaf node naturally inherits the semantic labels from the father nodes. Thus, each image possesses at least 2 category labels.
  This property enables Million-AID to be an aerial image dataset for hierarchical multi-label scene recognition. In total, there are 73 semantic labels contained in Million-AID. 
%   The number of scene images present in the dataset associated with each category label is listed in Table XX. To our knowledge, the number of scene categories in Million-AID is the largest compared to the existing publicly available RS scene classification datasets. 
  In this work, we treat the 51 fine-grained scenes as independently parallel categories for multi-class (single label) scene classification, and the 73 scene categories are employed for multi-label scene classification.
  
  \subsection{Dataset scale} 
\begin{figure}[!t]
      \centering
      \includegraphics[width=0.99\linewidth]
      {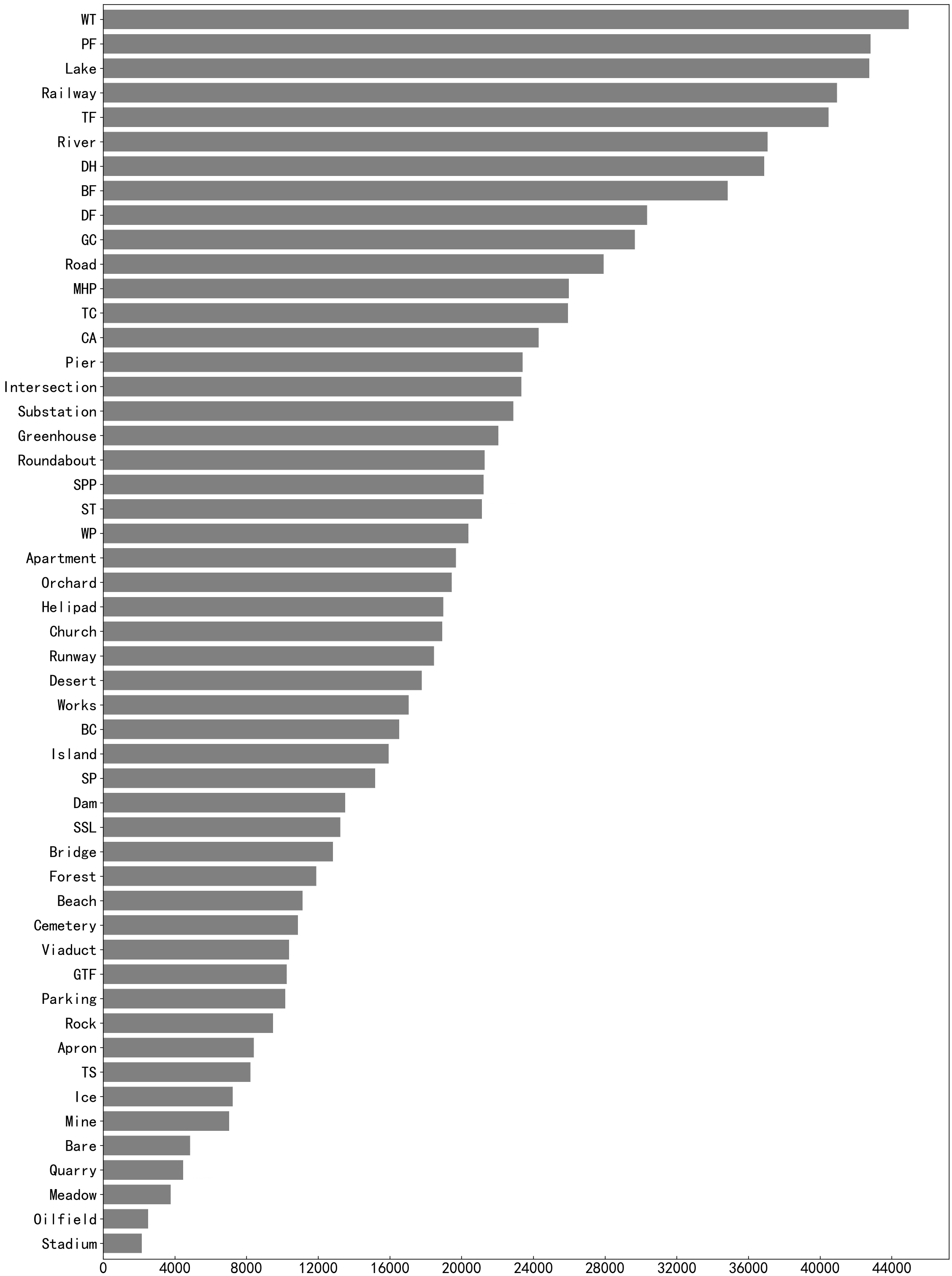}
      \caption{The number of instances in each scene category. Zoom for detail.}
      \label{figure:category_img_num}
      \vspace{-4mm}
  \end{figure}

  Apart from the wide coverage of semantic categories, Million-AID is characterized with a large scale. Particularly, the total number of images in Million-AID is 1,000,848. To the best of our knowledge, this is the first aerial scene classification dataset in which the number of images exceeds a million in the remote sensing community. As shown in Figure~\ref{figure:category_img_num}, the numbers of scene images vary greatly among different categories, endowing the dataset with the property of unbalanced distribution. Taking the widely used AID~\cite{AID} and NWPU-RESISC45~\cite{RESISC45} as a comparison, our proposed Million-AID surpasses them hugely in both the numbers of categories and images. Recently, data-driven methods particularly deep learning~\cite{deepreview,reichstein2019deep,cheng2020deepsc} have shown promising perspectives for intelligent aerial image interpretation, relying on the huge available dataset ontology. The Million-AID makes it possible to further boost the design and optimization of aerial scene interpretation algorithms using data-driven schemes.  
  
  \subsection{Geographical distribution} 
\begin{figure*}
      \centering
      \includegraphics[width=0.99\linewidth]
      {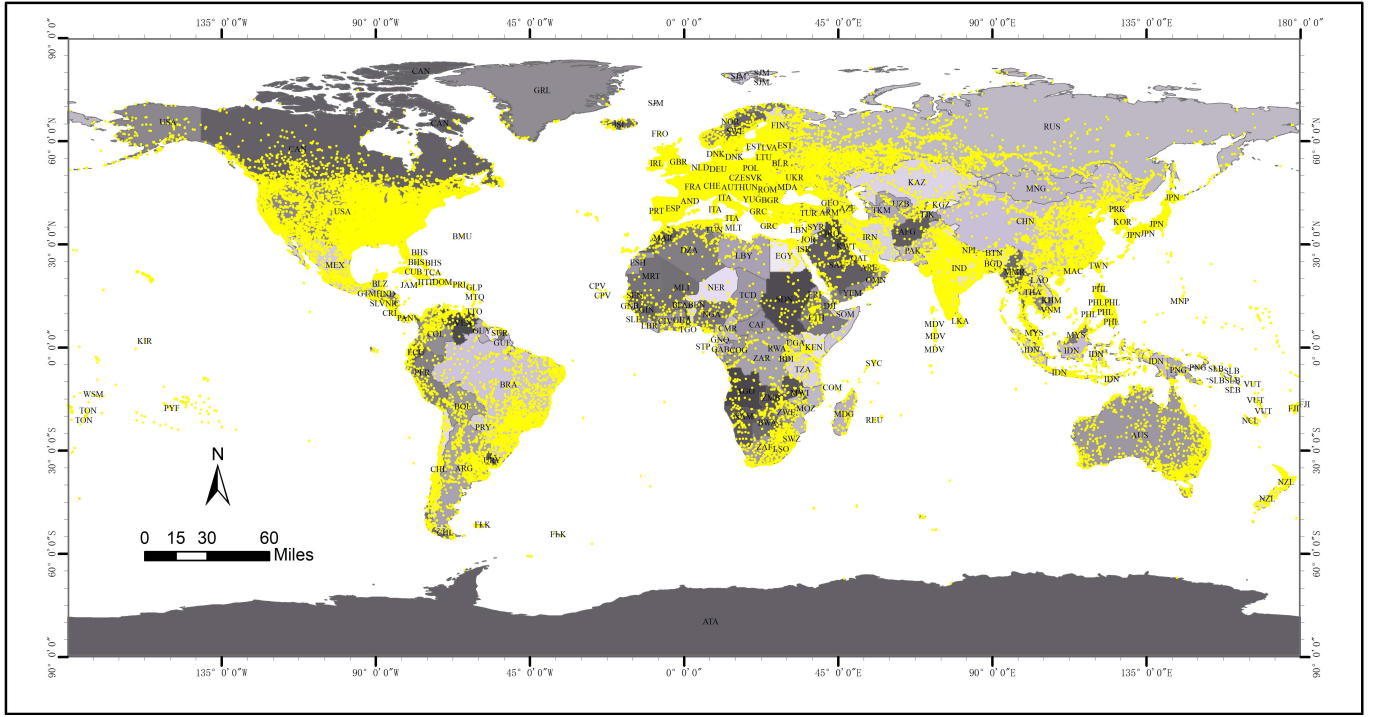}
      \caption{The distribution of scene images in Million-AID. To clearly express the image location and its distribution, more than a half of the total scene images are taken as the representative instances, and visually displayed using their geographical locations. The location of an aerial scene is represented by the central geographical coordinates of the image block.}
      \label{figure:img_distribution}
      \vspace{-4mm}
  \end{figure*}

  With the change of geographical environment, aerial scene images usually show different patterns in appearance, component, background. Hence, scenes in an aerial image dataset should be as widely distributed as possible to characterize their features in the real world. To this end, we collected the aerial scenes around the world by utilizing the geographical information as introduced in~\cite{Long2020DiRS}. The geographical distribution of aerial scenes in Million-AID is shown in Figure~\ref{figure:img_distribution}. It can be seen from the distribution map that the scene images are widely located all over the world. It is worth noting that most of the scene images are located on the land areas, and intensively distributed in cities or areas inhabited by humans. This is reasonable because it is in line with the reality that the semantic scenes of aerial images are usually closely associated with human production and living activities. 
  
  \subsection{Image variation}
  The rich variation of images can greatly enhance the diversity of a dataset, so as to better represent the scene and feature distribution in the real world. In Million-AID, the widths of scene images range from 100 to 30,000 pixels to approximate the scale variation of scenes in practical situations. The spatial resolution is 0.2m to 153m per pixel. As aerial imaging is easily affected by environmental factors, scene images in Million-AID are extracted under various circumstances,~\emph{i.e.}, viewpoint, weather, illumination, season, background, scale, resolution, geographical area,~\emph{etc}. These properties reflect the real challenges in the task of aerial scene recognition. 
%   These features ensure the completeness and authenticity of aerial scenes presented in the real world and also
  %   Image size distribution, resolution, imaging conditions, image sources, content complexity, ... structure, texture, color, season,  inter-class dissimilarity and intra-class similarity
  
  Furthermore, owing to the high complexity of ground features, scene content in aerial images usually show a remarkable difference in appearance characterized by various geometrical, structural, and textural attributes. This requires the created dataset with high intra-class diversity and inter-class similarity for developing interpretation algorithms with excellent generalization ability. The above-introduced properties and variation of scene images have provided sufficient assurance of intra-class diversity for Million-AID. Besides, scene images of sub-classes are typically contained in the same major classes. This enables the scene images of the sub-classes to possess properties of high inter-class similarity inherited from their common major classes. In general, the presented Million-AID is of great capacity to represent aerial scenes and feature distribution in the real world, and thus, facilitate the development of data-driven interpretation algorithms and the establishment of public comparison platforms.

\section{Scene Classification: A New Benchmark on Million-AID} \label{MultiClassSC}
 Data-driven algorithms represented by deep learning have been reported with overwhelming advantages over the conventional classification methods~\cite{AID,RESISC45}, and thus, dominated aerial image recognition in recent years~\cite{cheng2020deepsc}. In this section, we train a number of representative CNN models and conduct comprehensive evaluations for multi-class and multi-label scene classification on Million-AID, which we hope to provide a benchmark for future researches. 
 
 \subsection{Experimental setup}
  \textbf{Dataset partition:} 
  In order to make a comprehensive evaluation, the partition scheme is established for the baseline training and testing sets. Specifically, we extract training and test scene images located in different areas. In this configuration, we try to make the training and test data as spatially independent as possible. Consequently, there are 10, 000 scene images in the whole dataset of Million-AID randomly selected as the training subset, and the left images are fixed as the testing subset. Besides, the training set is characterized by long-tail distribution, which poses a great challenge to the scene classification model. 
 
\begin{table}[h!]
    \centering
    \caption{Summary of CNN models used in this work}
    \setlength{\tabcolsep}{3.0mm}{
    \begin{threeparttable}
    \begin{tabular}{c|c|c|c|c}
        \hline
        Model  &\#Layers  &\#Param. &Acc@1 (\%)  &Year\\ 
        \hline
        \hline
        AlexNet       &8    &60M   &56.52  &2012  \\
        VGG16         &16   &138M  &73.36  &2014  \\
        GoogleNet     &22   &6.8M  &69.78  &2014  \\
        ResNet101     &101  &44M   &77.37  &2015  \\
        DenseNet121   &121  &8M    &74.43  &2017  \\
        DenseNet169   &169  &14M   &75.60  &2017  \\
        \hline
    \end{tabular}
    \begin{tablenotes}
        \tiny
        \item\emph{Acc@1} indicates the Top-1 accuracy of CNN models tested on ImageNet. 
    \end{tablenotes}
    \end{threeparttable}}
    \label{tab:ModelDetail}
    \vspace{-3mm}
\end{table}
 
 \textbf{Model configuration: } 
  For image scene classification, the representative CNN models are employed for benchmarking experiments. Specifically, AlexNet~\cite{AlexNet}, VGG16~\cite{VGG}, GoogleNet~\cite{GoogleNet}, ResNet101~\cite{ResNet}, DenseNet121~\cite{DenseNet}, and DenseNet169~\cite{DenseNet} are selected to explore their scene classification performance on Million-AID. We chose these models in consideration of their broad applications in RS image interpretation, particularly in scene recognition. And it is apparent to observe that the employed CNN models consist of wide degrees of model depth and parameter scale, covering CNN frameworks from the shallow to deep ones, which can help to explore the classification performance more comprehensively and objectively. For the convenience of experimental implementation and fair performance comparison, we build a unified CNN library using PyTorch~\cite{PyTorch} for model training and testing. 
 
 \subsection{Multi-class scene classification}
%   \textbf{Training detail: }
 \subsubsection{Implementation detail} In~\cite{AID} and~\cite{RESISC45}, the aerial image features were directly extracted from the CNN models pretrained on ImageNet and then classified by support vector machines. By contrast, we deliver an end-to-end training scheme in which the~\emph{softmax} classifiers are integrated into the original CNN models.
 % Generally, we train the CNN models with parameters pretrained on ImageNet. the scratch to define our own network. Even learning the CNN models require massive samples, our training scheme is feasible owing to our large-scale training subsets as introduced before. Besides, in order to further explore the classification capability, we also conduct training strategy by fine-tuning the CNN models pretrained on ImageNet. 
 For efficient model adaption, we employ the training strategy by fine-tuning CNN models pretrained on ImageNet. For a fair comparison, we keep the training parameters consistent with different models. Specifically, the number of total iterations is set to be 50 epochs for sufficient parameter adaption considering the scalable training sets and stochastic gradient descent is utilized as the optimization strategy. The batch size is set to be 32. The initial learning rate is 0.01 and divided by 10 every 20 epochs. The weight decay and momentum are 0.005 and 0.9, respectively. The hardware is based on the Intel Xeon E5 CPU and the NVIDIA Tesla V100 GPU with 16GB memory.   

 \subsubsection{Evaluation protocols}
 For performance evaluation, we employ the commonly used overall accuracy (OA), average accuracy (AA), confusion matrix (CM), and Kappa coefficient (Kappa) to measure the classification results. The OA and CM are defined as the same as those in~\cite{AID,RESISC45}. Specifically, the OA is defined as the number of correctly predicted images divided by the total number of predicted images in the test dataset. The OA measures the classification performance on the whole dataset from a quantitative perspective while regardless of the classification performance on the single class. By contrast, AA is calculated by the mean value of classification accuracy of all classes. The CM can present the classification performance of a model on each class. Each row of a CM represents the actual instances in a predicted class while each column reveals the predicted instances in an actual class. The CM makes it convenient to explore a model's classification capability on the confusing classes. Kappa coefficient which can be calculated based on CM, is a robust measure since it takes into account the classification reliability for categorical items.

\subsubsection{Experimental results} 
 \textbf{Baseline results:} Table~\ref{tab:BaselineSingleLabel} illustrates the scene classification result from different CNN models. We can see that VGG16, GoogleNet, ResNet101, DenseNet121, and DenseNet169 achieve significantly better classification results when compared with AlexNet. Note that AlexNet is a shallow CNN framework with only 5 convolutional layers while the others possess more convolutional layers, which can extract highly abstract information for scene content representation. Thus, the deeper CNN models gains classification performance on OA, AA, and Kappa. This result demonstrates the superiority of the deep CNN frameworks, which is consistent with the classification of natural images~\cite{russakovsky2015imagenet}. 

 Particularly, VGG16 outperforms AlexNet and gives comparable results with some of the deeper models,~\emph{e.g.}, GoogleNet, ResNet101. This phenomenon stems from the advantage of the large scale of parameters possessed by VGG16 network. And the batch normalization operation incorporated in VGG16 network also helps to relieve the internal convariate shift problem~\cite{ioffe2015batch} reflected by the complex content of aerial images. Benefiting from the elaborately designed inception module, GoogleNet can gather features with different receptive fields in one layer, which makes it suitable for processing aerial scene images of high variation. 
% These properties enable the CNN models to learn excellent feature representation for diverse scene content and reveal their potential in aerial scene classification. 

\begin{table*}
\centering
\caption{Performance of single-label scene classification with different CNN models (\%)}
\setlength{\tabcolsep}{5.2mm}{
\begin{tabular}{c|c|c|c|c|c|c}
\hline
  Metric  &AlexNet  &VGG16  &GoogleNet  &ResNet101  &DenseNet121  &DenseNet169 \\ 
\hline
\hline
  OA      &67.53  &77.47  &77.37  &77.36  &79.04  &78.99 \\
  AA      &63.18  &74.58  &74.86  &74.58  &76.67  &76.67 \\
  Kappa   &66.61  &76.84  &76.73  &76.73  &78.46  &78.46 \\
\hline
\end{tabular}}
\label{tab:BaselineSingleLabel} 
% \vspace{-3mm}
\end{table*}

\begin{figure*}
    \centering
    \subfigure[]{
    \begin{minipage}[t]{0.31\linewidth}
    \centering
    \includegraphics[width=1\linewidth]{pictures/baseline/MAID_AN.pdf}
    % \caption{Caption}
    \end{minipage}
    }
    \subfigure[]{
    \begin{minipage}[t]{0.31\linewidth}
    \centering
    \includegraphics[width=1\linewidth]{pictures/baseline/MAID_VGG16BN.pdf}
    % \caption{Caption}
    \end{minipage}
    }
    \subfigure[]{
    \begin{minipage}[t]{0.31\linewidth}
    \centering
    \includegraphics[width=1\linewidth]{pictures/baseline/MAID_GN.pdf}
    % \caption{Caption}
    \end{minipage}
    }
    \quad
    \subfigure[]{
    \begin{minipage}[t]{0.31\linewidth}
    \centering
    \includegraphics[width=1\linewidth]{pictures/baseline/MAID_RN101.pdf}
    % \caption{Caption}
    \end{minipage}
    }
    \subfigure[]{
    \begin{minipage}[t]{0.31\linewidth}
    \centering
    \includegraphics[width=1\linewidth]{pictures/baseline/MAID_DN121.pdf}
    % \caption{Caption}
    \end{minipage}
    }
    \subfigure[]{
    \begin{minipage}[t]{0.31\linewidth}
    \centering
    \includegraphics[width=1\linewidth]{pictures/baseline/MAID_DN169.pdf}
    % \caption{Caption}
    \end{minipage}
    }
    \vspace{-1mm}
    \caption{Confusion matrix obtained by (a) AlexNet, (b) VGG16, (c) GoogleNet, (d) ResNet101, (e) DenseNet121, and (f) DenseNet169 on Million-AID dataset. Zoom for detail. }
    \label{fig:ConfusionMatrix_Baseline}
    \vspace{-4mm}
\end{figure*}

 Among the evaluated models, DenseNet121 and DenseNet169 outperform the others obviously. The densely connected nets can integrate features from different convolutional layers and thus enhance the representation ability of learned scene features. Note that DenseNet169 and achieves similar results with DenseNet121. This phenomenon reveals that a much deeper net would no longer bring performance improvement even with more dense connected layers. However, the OAs of all evaluated scene classification models are below 80\%. Therefore, more effective algorithms are expected to be developed toward semantic scene classification of aerial images. 

  \textbf{Analysis of different metrics:} When examining the performance by different metrics, we can find that Kappa and AA perform worse than OA. This is largely caused by the heavy unbalanced instance numbers of different scene categories. By referencing the confusion matrices as shown in Figure~\ref{fig:ConfusionMatrix_Baseline}, we can see that some categories with a relatively large number of scene images achieves high classification. For example,~\emph{wind turbine} and~\emph{river} contain over 44k and 37k instances, respectively. And the corresponding OAs achieved by DenseNet121 are close to 1. By contrast, some categories with a relatively small number of scene instances achieve lower classification accuracy. As a case in point,~\emph{stadium}, and~\emph{works} consist of only 2k and 17k instances while the corresponding OAs are only 0.49 and 0.37 by DenseNet121, respectively. As a result, the OA gains performance since it counts more on the total number of instances that are correctly classified while AA and Kappa are heavily influenced by the low accuracy of poorly classified categories. The difference is indicated by the AA as shown in Table~\ref{tab:BaselineSingleLabel}. Superficially, the unbalanced image numbers of scenes in Million-AID should be more in accordance with the scene distribution in the real world when compared with the existing scene classification datasets~\cite{RSSCN7,SIRI,RESISC45,PatternNet} in which each scene category share the same number of images. This implies that significant emphasis should be addressed to the property of category imbalance when developing algorithms for aerial scene classification. 
  
% \subsubsection{Confusion Matrix}
    \textbf{Confusion matrices:} By further investigating the confusion matrices (as shown in Figure~\ref{fig:ConfusionMatrix_Baseline}) of different CNN models, we can see that the deep CNN models, e.g., ResNet101 and DenseNet121/169, achieve much clearer confusion matrices than those of the shallow ones, e.g., AlexNet. It indicates that the deep CNN models have a better ability to distinguish different scene categories, which is consistent with the result from Table~\ref{tab:BaselineSingleLabel}. Several scene categories achieve classification accuracy approximate or equal to 1 as most of them show simple color, texture, and structure features in the scene images. Specifically, scene images like~\emph{desert} and~\emph{ice land} are mainly characterized with yellow and white components, respectively. The~\emph{terrace field} scene usually consists of distinct curve texture. In most cases, natural scenes like~\emph{river} and~\emph{sparse shrub land} show simple structure and monotonous content in the aerial images. Thus, these kinds of scenes can be easily distinguished from others benefiting from their highly recognizable features of image content. 

    Nevertheless, the majority of scene categories obtain the classification accuracy below 0.9 and quite a few categories obtain the classification accuracy below 0.5. Particularly, the~\emph{dry field} and~\emph{paddy field},~\emph{detached house} and~\emph{mobile home park} are heavily confused as they fall into similar land cover types, respectively. Many~\emph{stadium} images are misclassified as~\emph{ground track field} because of their high similarity of scene content. Most of the~\emph{beach} scenes are wrongly classified as~\emph{river} and~\emph{quarry} owing to their commonalities in structure and texture attributes. The same situation can also be observed between~\emph{dam} and~\emph{river} scenes. Notably, some scenes are easily misclassified as many different categories, such as~\emph{train station},~\emph{parking lot},~\emph{church}, and~\emph{works}. This phenomenon is mainly caused by the high intra-class variation of scene images that the algorithms cannot accurately distinguish them from each other. From this result, we can see that the Million-AID is a challenging dataset characterized with strong image variation of high inter-class similarity and intra-class diversity. Therefore, effective algorithms are desired to deal with these challenges, thereby, extracting excellent representations toward distinguishing different aerial scene categories.
% Although some scene categories can be easily distinguished from others and achieve classification accuracy approximate or equal to 1, most of them show simple color, texture, and structure features in the scene images. For the best result achieved by DenseNet121,  
\begin{table}[h!]
    \centering
    \caption{OA comparison among different datasets (\%)}
    \setlength{\tabcolsep}{4mm}{
    \begin{threeparttable}
        \begin{tabular}{c|c|c|c}
            \hline
            Dataset  &AlexNet  &VGG16  &GoogleNet \\ 
            \hline
            \hline
            AID  &86.86  &86.59 &83.44  \\
            AID*            &88.79  &93.72 &92.24  \\
            NWPU-RESISC45   &85.16  &90.36  &86.02  \\
            NWPU-RESISC45*  &87.19  &92.76   &91.71   \\
            Million-AID*    &67.53   &77.47  &77.37  \\
            % Million-AID     &76.84   &85.77  &85.28  \\
            \hline
        \end{tabular}
        \begin{tablenotes}
            \tiny
            \item\emph{AID*} indicates the average OAs of ten repeated experiments using our implemented CNN framework, so does the ~\emph{NWPU-RESISC45*} and~\emph{Million-AID}. The standard deviations are omitted since their negligible influence on the final result.
        \end{tablenotes}
    \end{threeparttable}}
    \label{table:BaselineCompare}
    \vspace{-3mm}
\end{table}

    \textbf{Comparison with existing benchmarks:} Many datasets have been established to promote the advancement of scene classification as detailed in~\cite{Long2020DiRS}. We compare the classification results of Million-AID with those of popular aerial scene classification datasets,~\emph{i.e.}, AID~\cite{AID} and NWPU-RESISC45~\cite{RESISC45}, considering their high quality and wide application. Table~\ref{table:BaselineCompare} describes the overall accuracy of different CNN models. The results show that our implemented CNN models (indicated with *) achieve better performance than that reported in the original publications, which confirms the rationality and superiority of our implemented framework and learning schemes. Thus, we are able to acquire reliable experimental results based on our established CNN library in this work. Obviously, scene classification on Million-AID reports significantly lower accuracy than that on AID and NWPU-RESISC45 with the utilized CNN models. This indicates Million-AID is a more challenging dataset than the compared ones. Note that the number of the testing images in Million-AID is dozens of times larger than that of other datasets. It means that a small decline of OA indicates a large amount of incorrectly classified scene images. Thus, Million-AID has the potential to serve as a reliable benchmark dataset for comprehensively evaluating and comparing the performance of different scene classification algorithms.

\subsection{Multi-label scene classification} \label{MultiLabelSC}
\begin{table*}
\centering
\caption{Performance of multi-label scene classification with different CNN models (\%)}
\setlength{\tabcolsep}{2.8mm}{
\begin{tabular}{c||ccc|ccc||ccc|ccc||c}
\hline
\multirow{2}{*}{Model} &\multicolumn{6}{c||}{$\tau=0.5$} &\multicolumn{6}{c||}{$\tau=0.75$} &\multirow{2}{*}{mAP}  \\ %[3pt]
\cline{2-13}
~ &CP  &CR  &CF1 &OP  &OR  &OF1  &CP  &CR  &CF1 &OP  &OR  &OF1 &~ \\ %[3pt] 
\hline
\hline
AlexNet       &71.45  &48.19  &57.56  &76.19  &62.84  &68.87  &78.89  &38.51  &51.76  &85.65  &53.03 &65.50  &61.76\\
VGG16         &82.26  &62.20  &70.84  &86.98  &75.31  &80.72  &84.61  &54.29  &66.14  &91.70  &69.37  &78.99  &79.13 \\
GoogleNet     &51.79  &33.99  &41.04  &88.50  &59.47  &71.14  &50.99  &23.76  &32.42  &94.90  &47.02  &62.89  &60.03 \\
ResNet101     &79.38  &59.67  &68.13  &88.74  &77.31  &82.63  &76.83  &51.56  &61.71  &93.05  &70.93  &80.50  &80.42 \\
DenseNet121   &79.09  &56.21  &65.71  &89.74  &75.10  &81.77  &76.36  &47.75  &58.76  &94.20  &67.72  &78.79 &78.94 \\
DenseNet169   &78.54  &61.92  &69.24  &88.50  &78.55  &83.23  &78.52  &55.10  &64.76  &92.66  &73.10  &81.72  &80.99 \\
% Million-AID     &76.84   &85.77  &85.28  \\
\hline
\end{tabular}}
\label{table:BaselineMultiLabel}
\vspace{-5mm}
\end{table*}

\subsubsection{Implementation detail}
We employ the aforementioned CNNs to evaluate the performance of multi-label scene classification on Million-AID. The predicted labels via the last fully connected layer are activated by~\emph{sigmoid} function and generate confidences for each of the semantic categories similar to~\cite{MLRSNET}. The binary cross-entropy is employed to measure the distance between the prediction and the true label (which is either 0 or 1). All CNN models are initialized with parameters pretrained on ImageNet. The training and testing subsets are the same a those for multi-class scene classification except for the labels that are extended according to the category organization system as shown in Figure~\ref{figure:MAID_Samples}. For the adaption of classification models, we transform the hierarchical multi-label scene classification problem into a traditional multi-class classification problem, where each of the nodes in the category system is regarded as a single label. In this configuration, the existing classification algorithms can be extended effortlessly for multi-label scene classification. 

\subsubsection{Evaluation protocols}
Precision and recall are employed as evaluation metrics. For each image, the predicted scene labels are considered as positive if the confidences are greater than a threshold $\tau$. Precision is defined as the fraction of correctly annotated labels with respect to generated labels. The recall is defined as the fraction of correctly annotated labels with respect to ground-truth labels. Following conventional settings~\cite{CNNRNN,chen2019multi,lin2021multilabel}, we calculate the per-class precision (CP), recall (CR), F1 (CF1) and overall precision (OP), recall (OR), F1 (OF1) for performance evaluation, where the average is calculated over all classes and all testing scene images, respectively. For a fair comparison, we also compute the mean average precision (mAP), which is the mean value of average precision per class. Generally, the CF1 and, OF1, and mAP are relatively more important evaluation metrics to reflect the comprehensive performance. 

\begin{figure*}
    \centering
    \quad
    \subfigure[]{
    \begin{minipage}[t]{0.99\linewidth}
    \centering
    \includegraphics[width=1\linewidth]{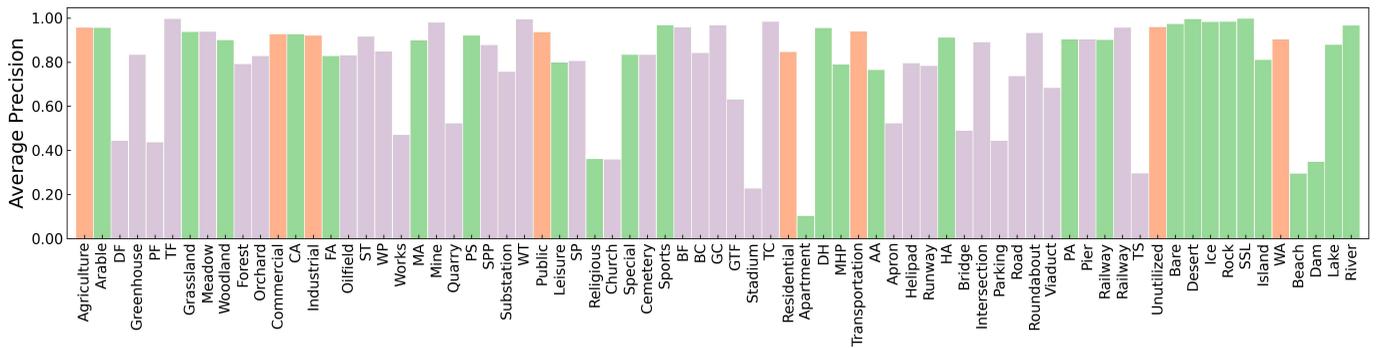}
    % \caption{Caption}
    \end{minipage}
    }
    \subfigure[]{
    \begin{minipage}[t]{0.99\linewidth}
    \centering
    \includegraphics[width=1\linewidth]{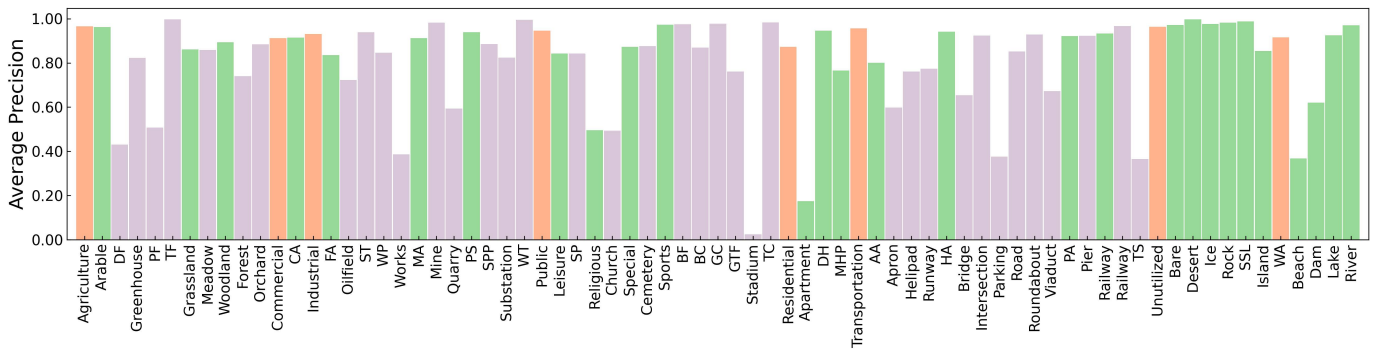}
    % \caption{Caption}
    \end{minipage}
    }
    \subfigure[]{
    \begin{minipage}[t]{0.99\linewidth}
    \centering
    \includegraphics[width=1\linewidth]{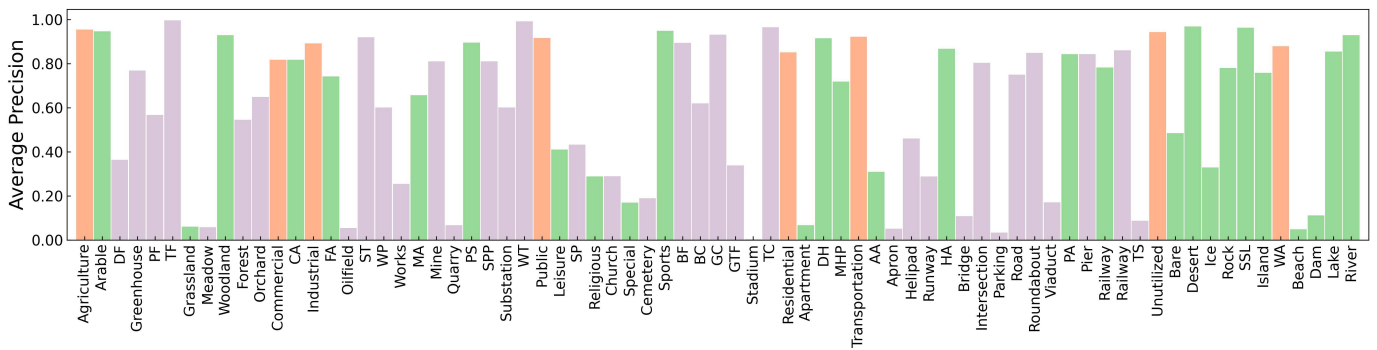}
    % \caption{Caption}
    \end{minipage}
    }
    \caption{Average precision results of (a) VGG16, (b) DenseNet169, and (c) GoogleNet on Million-AID dataset. The~\textcolor[rgb]{1., 0.6, 0.4}{orange} bars indicates predicted scene labels belonging to the first-level categories, the~\textcolor[rgb]{0.45, 0.8, 0.45}{green} bars the second-level categories, and the~\textcolor[rgb]{0.8, 0.7, 0.8}{plum} bars the third-level categories.}
    \label{fig:BaselineMultiLabel}
    \vspace{-4mm}
\end{figure*}

\subsubsection{Experimental results}
Quantitative results of multi-label scene classification on Million-AID are reported in Table~\ref{table:BaselineMultiLabel}. Obviously, the VGG16 model contains the most parameters while DenseNet169 has the most convolutional layers among the employed networks. As can be seen, VGG16 and DenseNet169 achieve comparable performance and obviously outperform the other networks. For per-class metrics, VGG16 obtains the CP of 82.26\% and CR of 62.20\%, achieving the best performance on CF1 of 70.84\% when~$\tau=0.5$. This result is slightly better than that from DenseNet169. Nevertheless, DenseNet169 achieves better performance on overall metrics, where the OP, OR, and OF1 are 88.50\%, 78.55\%, 83.23\%, respectively. Figure~\ref{fig:BaselineMultiLabel}(a) and (b) presents the average precision of each category when using VGG16 and DenseNet169, respectively. As can be seen, the two methods achieve similar classification performance for most categories. Superficially, the ResNet101 that shows superiority in both parameter scale and number of layers achieves the classification performance (mAP of 80.42\%) close to that of DenseNet169 (mAP of 80.99\%). However, when the scene images are assigned with the hierarchically multiple labels, the issue of data imbalance becomes prominent, which brings the problem of ``catastrophic forgetting''~\cite{goodfellow2013empirical,pfulb2018catastrophic}. As a result, the networks show weak performance on categories like~\emph{stadium} and~\emph{apartment}
% However, in comparison with VGG16, DenseNet169 shows insufficient ability in  recognizing~\emph{grassland},~\emph{meadow},~\emph{quarry},~\emph{ground track field},~\emph{stadium},~\emph{apron}, and~\emph{parking lot}. Consequently, DenseNet169 reports slightly worse performance on mAP than that of VGG16 for~$\tau=0.5$. 
When increasing $\tau$ to be 0.75, the OPs of all models gain significant improvement. This makes sense because the greater threshold value means the scene labels are predicted and filtered with higher confidence. However, all the recall metrics decline sharply, including the CR and OR metrics of different methods. As a result, the performance on CF1, OF1, and mAP metrics decline correspondingly in comparison with the result of which $\tau=0.5$. Nevertheless, DenseNet169 achieves the best performance on OF1 (81.72\%) and mAP (80.99\%), indicating its excellent ability in distinguishing different semantic scene categories.

An observation of interest is that there are shallow CNN models that significantly outperform the deep ones. A case in point is that for $\tau=0.5$ AlexNet achieves CF1 of 57.56\% and mAP of 61.76\%, which are 16.25\% and 1.73\% higher than those of GoogleNet, respectively. Even AlexNet achieves 2.27\% lower OF1 than that of GoogleNet, the former model shows superiority on OR. Besides, VGG16 reports the comprehensively better performance of mAP when compared with the deeper networks such as GoogleNet and DenseNet121. What is noteworthy is that the shallow networks,~\emph{i.e.}, AlexNet, and VGG16, contain particularly large-scale parameters compared with the others as detailed in Table~\ref{tab:ModelDetail}. With this superiority, the shallow networks are able to learn the relationship of scene categories at different levels of hierarchy. As a comparison, GoogleNet possesses more convolutional layers than those of AlexNet and VGG but with the least model parameters (6.8M). The experimental results show that GoogleNet provides the worst comprehensive performance (mAP of 60.03\%) among the employed CNN models. Taking the result of $\tau=0.5$ as an example, GoogleNet achieves CF1 of 41.04\% and mAP of 32.96\%, which are significantly poorer than those from other models. Simultaneously, the catastrophic forgetting problem becomes particularly prominent for GooogleNet. Figure~\ref{fig:BaselineMultiLabel}(c) presents the average precision of each category when using GoogleNet. As can be seen, many categories at the second and third semantic levels can not be recognized, resulting in poor CF1 and mAP. Therefore, GoogleNet shows relatively weak ability in learning the hierarchical relationship between different semantic scenes.   

Significantly, the biggest difference between VGG16 and AlexNet is that the former network possesses more convolutional layers and thus contains more than twice as many parameters as the former one. Hence, VGG16 gains remarkable improvement in classification performance. Although DenseNet121 consists of parameters at a scale comparable with that of GoogleNet, it possesses much more convolutional layers which help to significantly improve the performance of multi-label scene classification. As the depth of convolutional layers goes deeper, the performance improvement is also obvious,~\emph{i.e.}, the results from DenseNet121 and DenseNet169 as shown in Table~\ref{table:BaselineMultiLabel}. With the above analysis, it is natural to argue that both the parameter scale and depth of a CNN are crucial for recognizing the scene categories with multiple semantics. Intuitively, more parameters and deeper convolutional layers can enhance the network's ability to learn the heterogeneous characteristics of different scene categories, but also the ability to learn the homogeneous characteristics of scenes belonging to the same parent categories. This to some extent helps to reveal the hierarchical relationship between different semantic categories, which greatly improves the performance of hierarchical multi-label scene classification. Nevertheless, how to model the hierarchical rather than parallel relationships between different scene categories and further improve the performance of hierarchical multi-label scene classification remains to be further explored.  

\section{Transferring Knowledge From Million-AID} \label{KnowledgeTransfer}
Million-AID consists of large-scale aerial images that characterize diverse scenes. This provides Million-AID with rich semantic knowledge of scene content. Hence, it is natural for us to explore the potential to transfer the semantic knowledge in Million-AID to other domains. To this end, we consider two basic strategies,~\emph{i.e.}, fine-tuning pretrained networks for tile-level scene classification and hierarchical multi-task learning for pixel-wise semantic labeling.  

\subsection{Fine-tuning pretrained networks for scene classification} \label{sec:Finetune}
\subsubsection{Implementation detail}
A network trained from scratch is usually hard to capture the essential features of aerial scene content. Fine-tuning a pretrained CNN model has proven to be useful for aerial image interpretation~\cite{liu2018semantic,PSPNet2017,RESISC45,DOTA, DOTA2,Tong2019GID}, of which performance is improved by leveraging content knowledge from other domains. Particularly, CNN models are usually pretrained on natural image archives,~\emph{e.g.}, ImageNet~\cite{Imagenet}, and then fine-tuned on the target dataset for aerial image scene classification. The fine-tuning strategy has been regarded as a common solution to relieve the data scarcity problem for scene classification model adaption. Likewise, we employ the fine-tune learning strategy to verify the generalization ability of Million-AID dataset.

To verify the superiority of Million-AID, we first train CNN models for scene classification using all images in Million-AID. The CNN models pretrained on Million-AID are then fine-tuned with images in the target scene classification datasets,~\emph{i.e.}, AID~\cite{AID} and NWPU-RESISC45~\cite{RESISC45}. Similar to the dataset partition scheme in~\cite{AID,RESISC45}, 20\% images are randomly selected as training set and the rest 80\% as the test set. We repeat this operation ten times to reduce the influence of randomness and obtain reliable classification results. The epidemic CNN networks presented in Section~\ref{MultiClassSC} are employed to comprehensively evaluate the superiority of Million-AID. The learning rate is set to be 0.01 in the pretrain phase. In order to effectively utilize the scene knowledge learned from initial datasets, the learning rates in the fine-tuning phase are set to be 0.001 for all models. Through this step-wise optimization scheme, we are able to transfer the learned scene knowledge of Million-AID better to adapt to the target datasets. The other training parameters are set the same as those for multi-class scene classification in Section~\ref{MultiClassSC}. As a comparison, all the employed CNN networks are fine-tuned with the models pretrained on ImageNet. We also report the scene classification results from models trained from scratch, where the learning rates are also set to be 0.001 for consistency. The evaluation protocols are the same as those for multi-class scene classification as introduced in Section~\ref{MultiClassSC}. 

\begin{table*}
\centering
\caption{Classification accuracy (\%) on AID dataset using different Training schemes }
\begin{threeparttable}
\setlength{\tabcolsep}{3.2mm}{
\begin{tabular}{c|c|c|c|c|c|c|c}
\hline
Metric &Pretrain dataset  &AlexNet  &VGG16  &GoogleNet &ResNet101 &DenseNet121 &DenseNet169\\  
\hline
\hline
\multirow{3}{*}{OA}
~&W/O        &33.47 $\pm$ 2.15  &72.18 $\pm$ 0.49  &79.05 $\pm$ 0.89  &49.46 $\pm$ 2.07  &58.02 $\pm$ 0.74  &59.16 $\pm$ 0.52 \\
~&ImageNet   &88.79 $\pm$ 0.40  &93.72 $\pm$ 0.21  &92.24 $\pm$ 0.21  &94.52 $\pm$ 0.25  &94.68 $\pm$ 0.19  &94.76 $\pm$ 0.21 \\
~&Million-AID  &\textbf{90.70 $\pm$ 0.43}  &\textbf{95.33 $\pm$ 0.28}  &\textbf{94.55 $\pm$ 0.23}  &\textbf{95.40 $\pm$ 0.19}  &\textbf{95.22 $\pm$ 0.26}  &\textbf{95.24 $\pm$ 0.35} \\
\hline
\multirow{3}{*}{AA}
~&W/O        &33.85 $\pm$ 2.35  &72.16 $\pm$ 0.54  &78.88 $\pm$ 0.88  &49.29 $\pm$ 2.06  &57.88 $\pm$ 0.73  &59.04 $\pm$ 0.51 \\
~&ImageNet   &88.52 $\pm$ 0.39  &93.38 $\pm$ 0.22  &91.78 $\pm$ 0.23  &94.18 $\pm$ 0.29  &94.39 $\pm$ 0.21  &94.44 $\pm$ 0.22 \\
~&Million-AID  &\textbf{90.46 $\pm$ 0.45}  &\textbf{95.14 $\pm$ 0.27}  &\textbf{94.30 $\pm$ 0.23}  &\textbf{95.17 $\pm$ 0.19}  &\textbf{94.97 $\pm$ 0.26}  &\textbf{95.00 $\pm$ 0.38} \\
\hline
\multirow{3}{*}{Kappa}
~&W/O        &31.09 $\pm$ 2.24  &71.19 $\pm$ 0.51  &78.31 $\pm$ 0.92  &47.63 $\pm$ 2.15  &56.50 $\pm$ 0.76  &57.69 $\pm$ 0.53 \\
~&ImageNet   &88.39 $\pm$ 0.42  &93.49 $\pm$ 0.21  &91.96 $\pm$ 0.22  &94.32 $\pm$ 0.26  &94.49 $\pm$ 0.20  &94.57 $\pm$ 0.22 \\
~&Million-AID  &\textbf{90.37 $\pm$ 0.44}  &\textbf{95.17 $\pm$ 0.29}  &\textbf{94.35 $\pm$ 0.24}  &\textbf{95.24 $\pm$ 0.20}  &\textbf{95.05 $\pm$ 0.27}  &\textbf{95.07 $\pm$ 0.37} \\
\hline
\end{tabular}}
\begin{tablenotes}
\tiny
\item*~\emph{W/O} indicates the classification models are trained from scratch. 
\end{tablenotes}
\end{threeparttable}
\label{tab:TransferAID} 
\vspace{-3mm}
\end{table*}

\begin{table*}
\centering
\caption{Classification accuracy (\%) on NPWU-RESISC45 dataset using different training schemes }
\begin{threeparttable}
\setlength{\tabcolsep}{3.2mm}{
\begin{tabular}{c|c|c|c|c|c|c|c}
\hline
Metric &Pretrain dataset  &AlexNet  &VGG16  &GoogleNet &ResNet101 &DenseNet121 &DenseNet169\\  
\hline
\hline
\multirow{3}{*}{OA}
~&W/O        &37.92 $\pm$ 0.70  &73.19 $\pm$ 0.44  &81.77 $\pm$ 0.56  &58.82 $\pm$ 0.74  &63.35 $\pm$ 0.34  &64.51 $\pm$ 0.47 \\
~&ImageNet     &87.19 $\pm$ 0.26  &92.76 $\pm$ 0.18  &91.71 $\pm$ 0.25  &94.06 $\pm$ 0.16  &93.90 $\pm$ 0.19  &94.11 $\pm$ 0.20 \\
~&Million-AID  &\textbf{88.24 $\pm$ 0.21}  &\textbf{93.62 $\pm$ 0.20}  &\textbf{93.40 $\pm$ 0.23}  &\textbf{94.20 $\pm$ 0.16}  &\textbf{94.21 $\pm$ 0.20}  &\textbf{94.26 $\pm$ 0.21} \\
\hline
\multirow{3}{*}{AA}
~&W/O        &37.92 $\pm$ 0.70  &73.19 $\pm$ 0.44  &81.77 $\pm$ 0.56  &58.82 $\pm$ 0.74  &63.35 $\pm$ 0.34  &64.51 $\pm$ 0.47 \\
~&ImageNet     &87.19 $\pm$ 0.26  &92.76 $\pm$ 0.18  &91.71 $\pm$ 0.25  &94.06 $\pm$ 0.16  &93.90 $\pm$ 0.19  &94.11 $\pm$ 0.20 \\
~&Million-AID  &\textbf{88.24 $\pm$ 0.21}  &\textbf{93.62 $\pm$ 0.20}  &\textbf{93.40 $\pm$ 0.23}  &\textbf{94.20 $\pm$ 0.16}  &\textbf{94.21 $\pm$ 0.20}  &\textbf{94.26 $\pm$ 0.21} \\
\hline
\multirow{3}{*}{Kappa}
~&W/O        &36.51 $\pm$ 0.72  &72.59 $\pm$ 0.45  &81.36 $\pm$ 0.58  &57.89 $\pm$ 0.75  &62.51 $\pm$ 0.35  &63.70 $\pm$ 0.48 \\
~&ImageNet     &86.89 $\pm$ 0.21  &92.60 $\pm$ 0.19  &91.52 $\pm$ 0.26  &93.92 $\pm$ 0.17  &93.76 $\pm$ 0.19  &93.98 $\pm$ 0.20 \\
~&Million-AID  &\textbf{87.97 $\pm$ 0.21}  &\textbf{93.48 $\pm$ 0.20}  &\textbf{93.25 $\pm$ 0.24}  &\textbf{94.07 $\pm$ 0.16}  &\textbf{94.08 $\pm$ 0.20}  &\textbf{94.13 $\pm$ 0.21} \\
\hline
\end{tabular}}
\begin{tablenotes} 
\tiny
\item*~\emph{W/O} indicates the classification models are trained from scratch. 
\end{tablenotes}
\end{threeparttable}
\label{tab:TransferNWPU} 
\end{table*}
\begin{figure*}
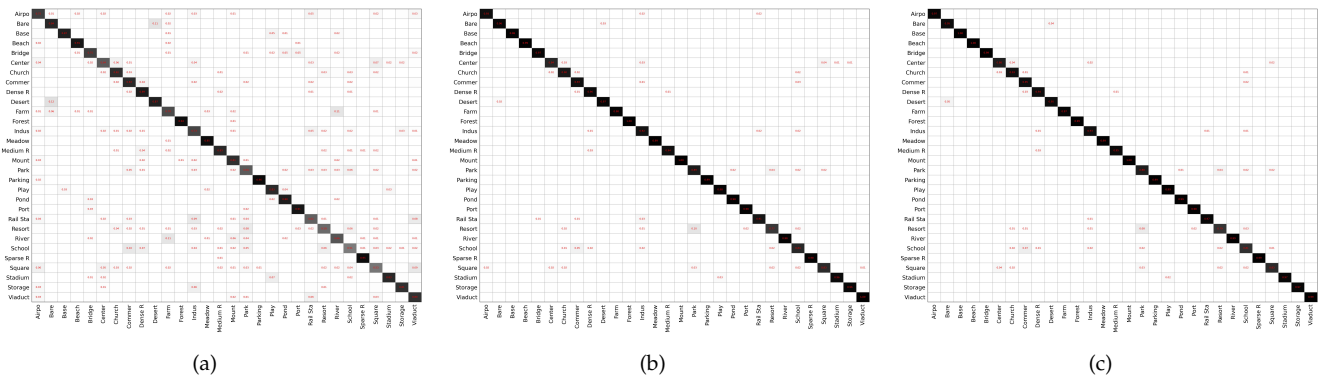

    \centering
    \subfigure[]{
    \begin{minipage}[t]{0.31\linewidth}
    \centering
    \includegraphics[width=1\linewidth]{pictures/finetune/AID_GN_SCRATCH.pdf}
    % \caption{Caption}
    \end{minipage}
    }
    \subfigure[]{
    \begin{minipage}[t]{0.31\linewidth}
    \centering
    \includegraphics[width=1\linewidth]{pictures/finetune/AID_DN169_IMNPRE.pdf}
    % \caption{Caption}
    \end{minipage}
    }
    \subfigure[]{
    \begin{minipage}[t]{0.31\linewidth}
    \centering
    \includegraphics[width=1\linewidth]{pictures/finetune/AID-RN101_MAIDPRE.pdf}
    % \caption{Caption}
    \end{minipage}
    }
    \caption{Confusion matrix obtained by (a) GoogleNet trained from scratch, (b) DenseNet169 pretrained on ImageNet, and (c) ResNet101 pretrained on Million-AID. Results are based on AID dataset. Zoom for detail.}
    \label{fig:ConfusionMatrixAID_Finetune}
\end{figure*}

\begin{figure*}
    \centering
    \subfigure[]{
    \begin{minipage}[t]{0.31\linewidth}
    \centering
    \includegraphics[width=1\linewidth]{pictures/finetune/RESISC45_GN_SCRATCH.pdf}
    % \caption{Caption}
    \end{minipage}
    }
    \subfigure[]{
    \begin{minipage}[t]{0.31\linewidth}
    \centering
    \includegraphics[width=1\linewidth]{pictures/finetune/RESCSC45_DN169_IMNPRE.pdf}
    % \caption{Caption}
    \end{minipage}
    }
    \subfigure[]{
    \begin{minipage}[t]{0.31\linewidth}
    \centering
    \includegraphics[width=1\linewidth]{pictures/finetune/RESISC45_DN169_MAIDPRE.pdf}
    % \caption{Caption}
    \end{minipage}
    }
    \caption{Confusion matrix obtained by (a) GoogleNet trained from scratch, (b) DenseNet169 pretrained on ImageNet, and (c) DenseNet169 pretrained on Million-AID. Results are based on NWPU-RESISC45 dataset. Zoom for detail.}
    \label{fig:ConfusionMatrixRESISC45_Finetune}
\end{figure*}

 \begin{figure*}
    \centering
    \includegraphics[width=0.99\linewidth]{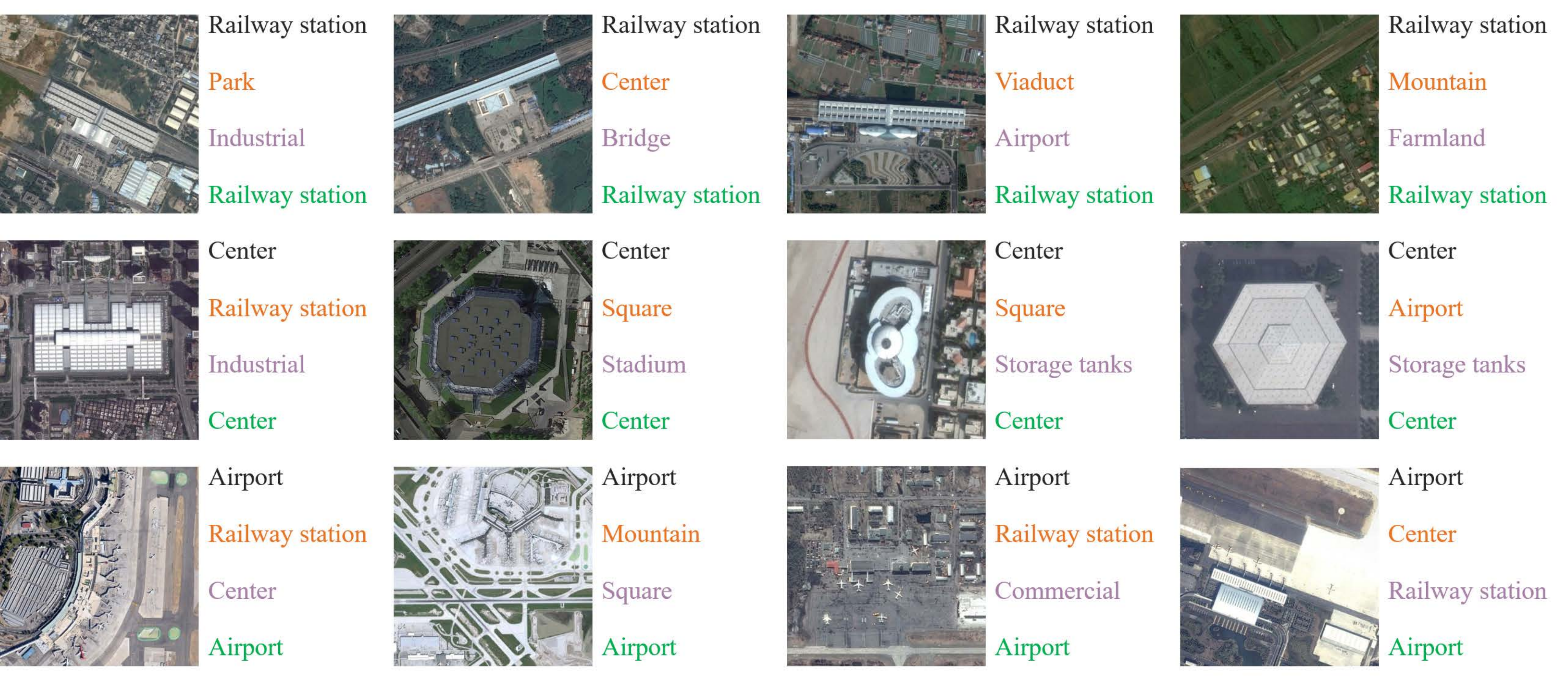}
    \vspace{-1mm}
    \caption{Example images and predictions on AID. For each training scheme, the model with the best classification result is selected for comparison. The black labels are the ground truth. The~\textcolor[rgb]{1., 0.6, 0.4}{orange} labels indicate predictions by GoogleNet trained from scratch, the~\textcolor[rgb]{0.8, 0.7, 0.8}{plum} labels the predictions by DenseNet169 pretrained on ImageNet, and the~\textcolor[rgb]{0.0, 0.58, 0.30}{green} labels the predictions by ResNet101 pretrained on Million-AID.}
    \label{fig:CompareResultAID_Finetune}
    \vspace{-3mm}
 \end{figure*}
 \begin{figure*}
    \centering
    \includegraphics[width=0.99\linewidth]{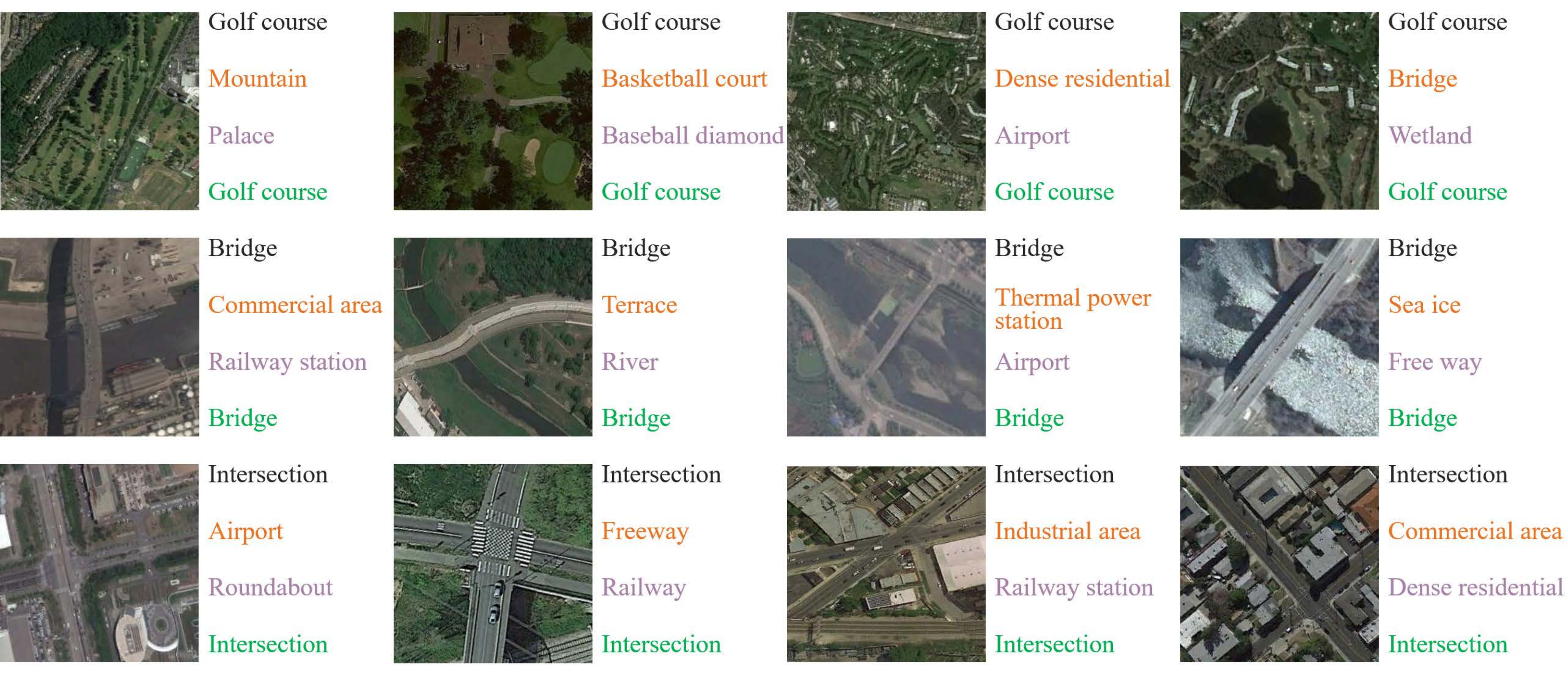}
    \vspace{-1mm}
    \caption{Example images and predictions on NWPU-RESISC45. For each training scheme, the model with the best classification result is selected for comparison. The black labels are the ground truth. The~\textcolor[rgb]{1., 0.6, 0.4}{orange} labels indicate predictions by GoogleNet trained from scratch, the~\textcolor[rgb]{0.8, 0.7, 0.8}{plum} labels the predictions by DenseNet169 pretrained on ImageNet, and the~\textcolor[rgb]{0.0, 0.58, 0.30}{green} labels the predictions by DenseNet169 pretrained on Million-AID.}
    \label{fig:CompareResultRESISC45_Finetune}
    % \vspace{-4mm}
 \end{figure*}

\subsubsection{Experimental results}
 Tables~\ref{tab:TransferAID} and~\ref{tab:TransferNWPU} illustrate the means and standard deviations of OA, AA, and Kappa on AID and NWPU-RESISC45, respectively. For each model, the best performance among different learning schemes (\emph{i.e.}, the models trained from scratch, fine-tuned on ImageNet, and fine-tuned on Million-AID) is reported in bold. By analyzing the tables, one can see that learning directly from scratch achieves the worst result. It indicates that optimizing CNN models for aerial scene classification can be difficult owing to the scarcity of training data and complexity of scene content. Thus, researchers often resort to extracting aerial scene features by models adapted well on natural images and then recognize aerial scenes by utilizing feature classifiers (\emph{e.g.}, SVM~\cite{AID,RESISC45,toward2017nogueira}). Compared with the models trained from scratch, the models pretrained on ImageNet and Million-AID can significantly improve the classification performance. 
%  As an example, fine-tuning pretrained AlexNet on NWPU-RESISC45 provides about 50\% higher accuracy, compared with the scratch learning strategy. Likewise, over 55\% higher accuracy can be achieved when fine-tuning pretrained AlexNet on AID. The other pretrained models also show a remarkable improvement of classification performance as shown in Tables~\ref{tab:TransferAID} and~\ref{tab:TransferNWPU}. 
 Figures~\ref{fig:ConfusionMatrixAID_Finetune} and~\ref{fig:ConfusionMatrixRESISC45_Finetune} provide the confusion matrices of the best results obtained by different learning schemes on AID and NWPU-RESISC45, respectively. It is shown that the classification performance of each scene category is significantly improved by the fine-tuned models. This confirms the importance of parameter initialization for CNN model adaption. In particular, it strongly demonstrates the effectiveness and positive significance of Million-AID for training CNN models toward aerial image scene classification. 
 
 An important observation is that all models pretrained on Million-AID achieve obviously better performance compared with those pretrained on ImageNet. Specifically, for both AID and NWPU-RESISC45, each considering CNN model pretrained on Million-AID provides the maximum accuracy. As shown in Figure~\ref{fig:ConfusionMatrixAID_Finetune} (b) and (c), by employing Million-AID for model pretraining, the classification accuracy of~\emph{railway station}, ~\emph{center}, and~\emph{airport} in AID reach 97\%, 88\%, and 97\%, which are 6\%, 4\%, and 4\% higher than using ImageNet, respectively. Likewise, impressive accuracy improvement can also be observed for scene categories in NWPU-RESISC45, such as~\emph{golf course},~\emph{bridge}, and~\emph{intersection} as shown in Figure~\ref{fig:ConfusionMatrixRESISC45_Finetune} (b) and (c). Figures~\ref{fig:CompareResultAID_Finetune} and~\ref{fig:CompareResultRESISC45_Finetune} provide the corresponding example images and predictions on AID and NWPU-RESISC45, respectively. It is shown that the models pretrained on Million-AID can better distinguish between semantic scenes with similar characteristics. 
%  Particularly, by employing Million-AID for model pretraining, the classification accuracy of~\emph{railway station},~\emph{church}, ~\emph{center}, and~\emph{airport} in AID reach 97\%, 93\%, 88\%, and 97\%, which are 6\%, 3\%, 4\%, and 4\% higher than using ImageNet, respectively.  
%  When it comes to the AID dataset, only the RestNet101 model pretrained on ImageNet reports slight accuracy improvement compared to that from Million-AID pretrained model. Nevertheless, the Million-AID pretrained model presents much lower standard deviations, which indicates its advantage in stability. 
 Intuitively, due to the difference in spatial pattern, texture structure, and visual appearance, there are a gigantic feature and semantic discrepancies between the natural and aerial image content. Hence, the CNN models pretrained with natural images may not be generally applicable to reduce this gap for aerial image interpretation. By contrast, the models trained with pure large-scale aerial images can naturally grasp the unique characteristics and knowledge of image content. With this advantage, the subsequent CNN models fine-tuned with aerial images in the target datasets are able to learn better features for aerial scene content representation, and thus, outperform those using the natural images. 
 
From shallow networks (\emph{e.g.}, AlexNet, VGG16, and GoogleNet) to the deeper ones (\emph{e.g.}, ResNet101, DenseNet121, and DenseNet169), the performance difference between ImageNet and Million-AID pretrained models become smaller. As an example on NWPU-RESISC45, AlexNet, VGG16, and GoogleNet pretrained on Million-AID achieves 1.05\%, 0.86\%, and 1.69\% higher OAs than the results from ImageNet pretrained models, respectively. This accuracy difference decreases to 0.14\%, 0.31\%, and 0.15\% when it comes to ResNet101, DenseNet121, and DenseNet169, respectively. The results on AID also show a similar phenomenon. This makes sense because the deeper the network, the more likely the learned features adapted to the target aerial images, resulting in comparable performance among different learning strategies. Nevertheless, the models pretrained with Million-AID still show superiority, which confirms the strong generalization ability of Million-AID. To our knowledge, it is the first time to be observed that CNN models pretrained with pure large-scale aerial images are verified to surpass those using natural images. Prior to this, many CNN models are usually pretrained using the ImageNet dataset and then fine-tuned on the target dataset for aerial image scene classification owing to the lack of available large-scale aerial image archives. With the above observation and results, it is natural to argue that the proposed Million-AID can make an advancement for the use of CNN models in aerial image scene classification, opening up a promising direction to support parameter initialization of CNN models toward various aerial image interpretation tasks such as object detection and semantic segmentation. 

\begin{figure*}
    \centering
    \includegraphics[width=0.85\linewidth]{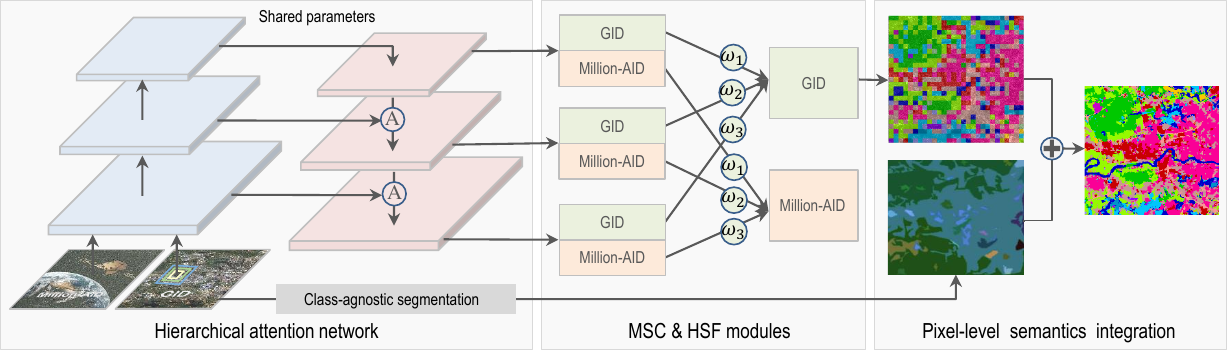}
    \caption{The framework of hierarchical multi-task learning for pixel-wise semantic labeling. Each pair of framed~\emph{GID} and~\emph{Million-AID} denotes the multi-task classification branches for GID and Million-AID, respectively. The sign of~\textcircled{A} indicates the attention module and the~\textcircled{+} the majority voting process.}
    \vspace{-4mm}
    \label{fig:HMTL}
\end{figure*}

\subsection{Hierarchical multi-task learning for semantic labeling} \label{HMLM}
\subsubsection{Method of overview}
The conventional CNN learns scene features via stacked convolutional layers and the output of the last fully connected layer is usually employed for scene representation. However, learning stable features from a single layer can be a difficult task because of the complexity of scene content. 
% The features from different convolutional layers can learn valuable information of objects, texture structures, spatial patterns, and semantic clues associated with specific scene recognition. However, due to the complexity of aerial scene content, features from a single scale are not able to assemble the rich information of features from multiple layers.
Moreover, data sparsity which is a long-standing notorious problem can easily lead to model overfitting and weak generalization ability because of the insufficient knowledge captured from limited training data. To relieve the above issues, we introduce a hierarchical multi-task learning method and further explore how well the knowledge contained in Million-AID can be transferred to boost the pixel-wise semantic parsing of aerial images. To this end, the GID~\cite{Tong2019GID}, which consists of a training set with tile-level scenes and large-size test images with pixel-wise annotations, has provided us an opportunity to bridge the tile-level scene classification toward pixel-wise semantic labeling. Generally, the presented framework consists of four components,~\emph{e.g.}, hierarchical attention network, multi-task scene classification, hierarchical semantic fusion, and pixel-wise semantics integration, as shown in Figure~\ref{fig:HMTL}. 
% Actually, convolutional features from different layers can capture rich information of objects, texture structures, and spatial patterns which are essential for specific scene recognition. 
% Due to the complexity of aerial scene content, the scale information of objects, texture structures, and spatial patterns associated with specific scene semantics are usually provided    
% With the change of imaging condition and scene content, the scales of objects, texture structures, and spatial patterns associated with scene semantics can vary dramatically.
% Thus, the feature from a single scale usually shows the insufficient capacity for the representation of diverse scene content.

\begin{figure}[!ht]
        \centering
        \includegraphics[width=0.75\linewidth]{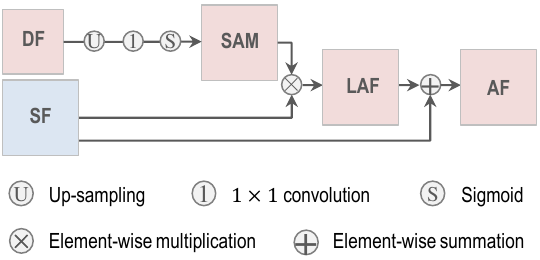}
        \vspace{-2mm}
        \caption{Attention module in hierarchical attention network.}
        \label{fig:HAN}
        \vspace{-4mm}
    \end{figure}

 \textbf{Hierarchical attention network (HAN):} The high-resolution features from shallow convolutional layers can learn valuable visual information of small objects, texture structures, and spatial patterns associated closely with specific scene content while the semantic clues is insufficient. To compensate for this defect, the hierarchical attention features are learned via transmitting the semantic information from the deep layers to the shallower ones inspired by~\cite{lin2017feature,Backward2019}. Specifically, the deep-layer feature (\textbf{DF}) is firstly upsampled to the same spatial size as the shallow-layer feature (\textbf{SF}) to maintain the semantic information as much as possible. The upsampled semantic feature is processed by a $1\times1$ convolutional layer for dimension reduction. And the sigmoid function is then employed to generate the semantic attention map (\textbf{SAM}), which possesses the same channel number as the~\textbf{SF}. Element-wise multiplication is conducted between the~\textbf{SF} and~\textbf{SAM} to generate local attention feature (\textbf{LAF}). Finally, the~\textbf{SF} and~\textbf{LAF} are assembled by element-wise summation and output the final attention feature (\textbf{AF}). The whole process is illustrated in Figure~\ref{fig:HAN}. By repeatably transmitting the deep-layer features to the shallower convolutional layers, we are able to construct the hierarchical attention network (as shown in Figure~\ref{fig:HMTL}) that incorporates semantic and visual information for scene representation. 

 \textbf{Multi-task scene classification (MSC):} With the hierarchical attention features, a semantic scene can be represented using the features of multiple scales. As shown in Figure~\ref{fig:HMTL}, three streams are constructed for multi-scale feature extraction. For each stream, the hierarchical attention feature is further processed by global average pooling and generates the feature for scene representation. Then, the multi-task classification branches are designed to recognize scenes from different datasets. 
 %   As an inductive transfer learning paradigm, multi-task learning aims at achieving high performance in the target task by training model for multiple relevant tasks~\cite{TL2010,MLT2021}. 
 By learning the sharing parameters for different tasks, multi-task learning enables the knowledge learned in one task to be utilized in the others, and thus, improve the generalization ability of a trained model~\cite{TL2010,MLT2021}. The presented Million-AID is composed of rich semantic scenes and massive instances that characterize the land cover features. It is reasonable to transfer the land cover knowledge contained in Million-AID to boost the semantic classification of aerial images. With this in mind, the multi-task learning is conducted on Million-AID and the challenging GID~\cite{Tong2019GID}, which is established for pixel-wise semantic classification of land cover. For simplicity, two branches at each of the hierarchical output layers are designated for the scene-level classification of images in Million-AID and GID, respectively. The summation of weighted losses is used for model adaption:
 \begin{align}
        Loss^g &= \sum_{s=1}^{S} w_s \textrm{CE}_s^g, \\
        Loss^m &= \sum_{s=1}^{S} w_s \textrm{CE}_s^m, \\
        Loss &= \mu_g Loss^g + \mu_m Loss^m,
\end{align}
 where $CE_s^g$ represents the cross entropy loss of scene classification using the image features of scale $s \in {1, 2, ..., S}$ for GID while $CE_s^m$ for Million-AID. $w_s$ indicates the loss weight for the classification at scale $s$. $\mu_g$ and $\mu_m$ (where $\mu_g + \mu_m = 1$) indicate the weighted importance of different tasks,~\emph{i.e.}, scene classification on GID and Million-AID, respectively. In this work, we aim at improving the classification performance on GID by knowledge transfer from Million-AID. Hence, the semantic classification on GID is regarded as the main task while the scene classification on Million-AID serves as the auxiliary task~\cite{ruder2017overview} to still reap the benefits of multi-task learning strategy. 
 
 \textbf{Hierarchical semantic fusion (HSF): } To give full play of the advantages of hierarchical attention features, the classification results with different feature scales are integrated. Using the feature at scale $s$, the classification probability vector $p_s(I)$ of image $I$ is obtained by a~\emph{softmax} layer: 
  \begin{equation}
     p_s(I) = \{p_{s, 1}(I),\ p_{s, 2}(I),\ ...,\ p_{s, N}(I)\},\ \ p_s(I) \in \mathbb{R}^N
  \end{equation}
 where $p_{s, n}$ represents the probability that $I$ belongs to class $n$ using the feature of scale $s$. Essentially, the predictions at different scales reflect the probability that a classified scene belongs to individual categories from the perspective of different feature levels. Hence, it is reasonable to integrate the predictions of different scales. To this end, a summation of weighted probabilities is performed as the final prediction:
 \begin{equation}
     \widehat{p}_n(I) = \frac{\sum_{s=1}^{S} w_s p_{s, n}(I)}{\sum_{s=1}^{S} w_s} 
 \end{equation}
 where $\widehat{p}_n(I) \in [0, 1]$ indicates the probability that image $I$ belongs to class $n$. $w_s$ represents the weight for scale $s$, which serves as the loss weight for the corresponding classification stream. The integration of weighted probabilities aims to provide a more stable prediction result. Then the predicted scene category of image $I$ is expressed as: 
  \begin{equation}
    l(I) = \mathop{\arg\max}_{n \in [1, 2, ..., n]} \ \widehat{p}_n(I)
  \end{equation}
  where $l(I)$ is the category label of image $I$. 
  
  \begin{table*}[h]
        \centering
        \caption{Weights influence on different tasks}
        \setlength{\tabcolsep}{3.5mm}{
        \begin{tabular}{cc|ccc||ccc}
            \hline
            \multirow{2}{*}{$\mu_g$} &\multirow{2}{*}{$\mu_m$} &\multicolumn{3}{c||}{GID} &\multicolumn{3}{c}{Million-AID} \\
            \cline{3-8}
            ~  &~  &Kappa (\%)  &OA (\%)  &mIoU (\%) &Kappa (\%)  &OA (\%)  &AA (\%) \\
            \hline
            \hline
            %  0.1  &0.9      &59.00    &69.45    \\
             0.1  &0.9     &62.85   &69.06    &39.88 &\textbf{90.36} &\textbf{90.62} &\textbf{89.55}  \\
            %  0.3  &0.7     &61.94     &71.46   \\
             0.3  &0.7     &65.15   &71.00    &41.85 &89.44 &89.72 &88.91  \\
            %  0.5  &0.5     &\textbf{63.44}  &\textbf{72.76} \\
             0.5  &0.5    &\textbf{66.65} &\textbf{72.38}  &\textbf{42.71} &\textbf{\textcolor[rgb]{0.1, 0.1, 0.1}{89.67}} &\textbf{\textcolor[rgb]{0.1, 0.1, 0.1}{89.94}} &\textbf{\textcolor[rgb]{0.1, 0.1, 0.1}{89.14}} \\
            %  0.7  &0.3     &62.80     &72.47      \\
             0.7  &0.3    &66.14    &72.02   &41.75  &88.98  &89.27  &87.84 \\
            \hline
        \end{tabular}}
        \label{tab:AblationWeight} 
  \end{table*}
\begin{figure*}
    \centering
    \subfigure[]{
    \begin{minipage}[t]{0.23\linewidth}
    \centering
    \includegraphics[width=1\linewidth]{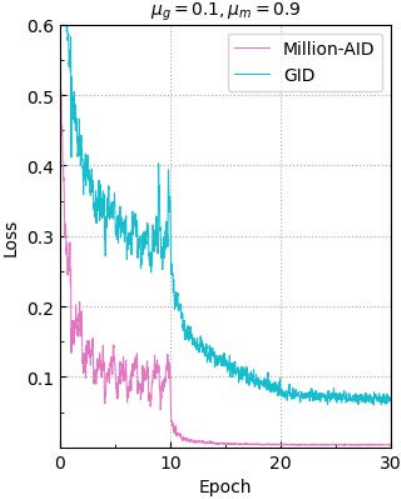}
    % \caption{Caption}
    \end{minipage}
    }\vspace{-0.5mm}
    \subfigure[]{
    \begin{minipage}[t]{0.23\linewidth}
    \centering
    \includegraphics[width=1\linewidth]{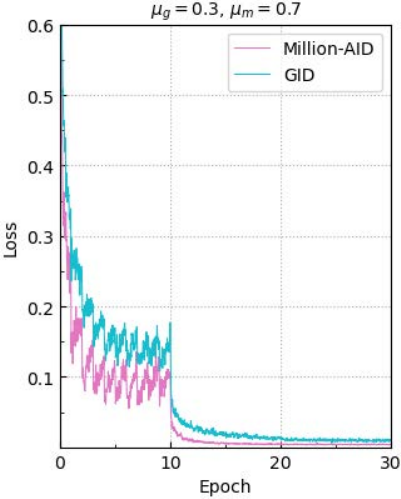}
    % \caption{Caption}
    \end{minipage}
    }\vspace{-0.5mm}
    % \quad
    \subfigure[]{
    \begin{minipage}[t]{0.23\linewidth}
    \centering
    \includegraphics[width=1\linewidth]{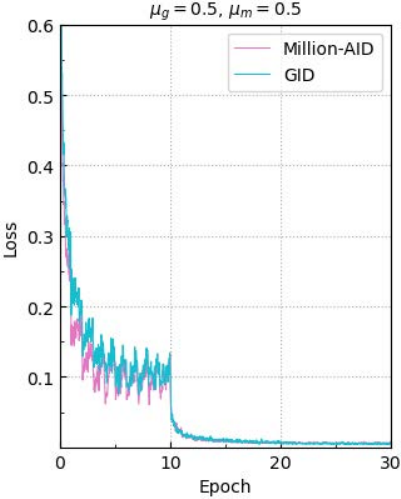}
    % \caption{Caption}
    \end{minipage}
    }\vspace{-0.5mm}
    \subfigure[]{
    \begin{minipage}[t]{0.23\linewidth}
    \centering
    \includegraphics[width=1\linewidth]{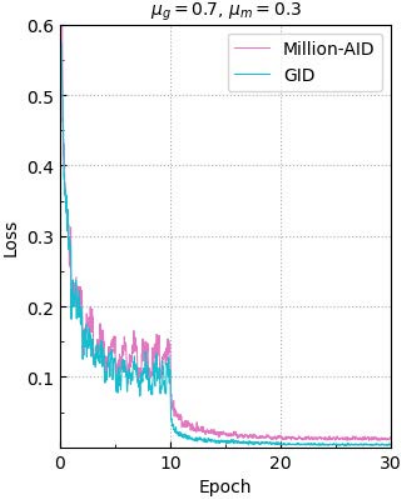}
    % \caption{Caption}
    \end{minipage}
    }\vspace{-0.5mm}
    \caption{Training loss with respect to different setups of learning weights. The models converge successfully for both classification tasks with different learning weights, which confirms the validity of the learned classification model.}
    \label{fig:MTLossAblation}
    \vspace{-4mm}
\end{figure*}
  
  \textbf{Pixel-wise semantics integration (PSI):} With the above procedures, we can obtain the semantic grid map by tile-level classification for interpreting a large-size aerial image. Here, each semantic grid corresponds to a tile-level classification result. For more accurate results with pixel-wise semantics, the semantic boundaries in an aerial image appear the great importance. We therefore perform class-agnostic image segmentation to produce accurate semantic boundaries. For simplicity, we employ object-based segmentation and a majority voting strategy to generate pixel-wise semantic labeling result. Specifically, the selective search algorithm~\cite{van2011segmentation} is conducted on the raw aerial image and produces a homogeneous segmentation map. Let $r$ be a homogeneous region in the segmentation map. The majority voting algorithm is then performed to determine the semantic class of $r$ denoted as $l(r)$ by referencing the semantic grid map: 
   \begin{equation}
    l(r) = \mathop{\arg\max}_{n \in [1, 2, ..., n]} \ |r_n|
   \end{equation}
  where $|r_n|$ denotes the number of pixels enclosed in $r$ and labeled as class $n$ in the semantic grid map. To enhance the discrimination ability of scene content, we represent the scene with multi-scale context, of which semantic meaning is identified by the central point as illustrated in Figure~\ref{fig:header}(d). By integrating multi-scale contextual information with tile-level classification, the whole pipeline falls into a hybrid classification framework~\cite{Tong2019GID}. Nevertheless, our method focuses more on improving the tile-level scene classification performance by utilizing hierarchical attention features and transferring aerial scene knowledge from the relevant domain,~{\em i.e.}, Million-AID.  
  
  \subsubsection{Implementation detail} ResNet50~\cite{ResNet} is employed as the backbone, where the last three residual blocks of conv3\_x, cov4\_x, and conv5\_x are utilized to extract hierarchical attention features in three streams. The loss weights for the three streams are empirical set to be $w_1=0.25$, $w_2=0.5$, and $w_3=1.0$ according to the ratios of channel numbers of the hierarchical layers, respectively. The~{\em fine classification set} of GID~\cite{Tong2019GID}, which contains 15 challenging semantic categories, is employed for performance evaluation. Specifically, the proposed model is trained with the subset of 30k tile-level scene patches and then tested on the subset of 10 Gaofen-2 images (6800$\times$7200) with pixel-wise semantic labels. The multi-scale contextual information is extracted with the windows of 56$\times$56, 112$\times$112, and 224$\times$224, which are consistent with the sizes of training samples. As training samples in GID are highly overlapped, image flipping and rotation (90$^\circ$, 180$^\circ$, and 270$^\circ$) are conducted for data augmentation, resulting in 120k tile-level scene samples. Correspondingly, 120k scene images in Million-AID are randomly selected as the training set, according to the ratios of instance number of each scene category. Training parameters are set the same as those for multi-class scene classification in Section~\ref{MultiClassSC} except the number of iteration set as 30 epochs. The OA, Kappa, and mean-Intersection-over-Union (mIoU)~\cite{minaee2021image} are employed for performance evaluation.

\begin{figure*}
    \centering
    \includegraphics[width=1\linewidth]{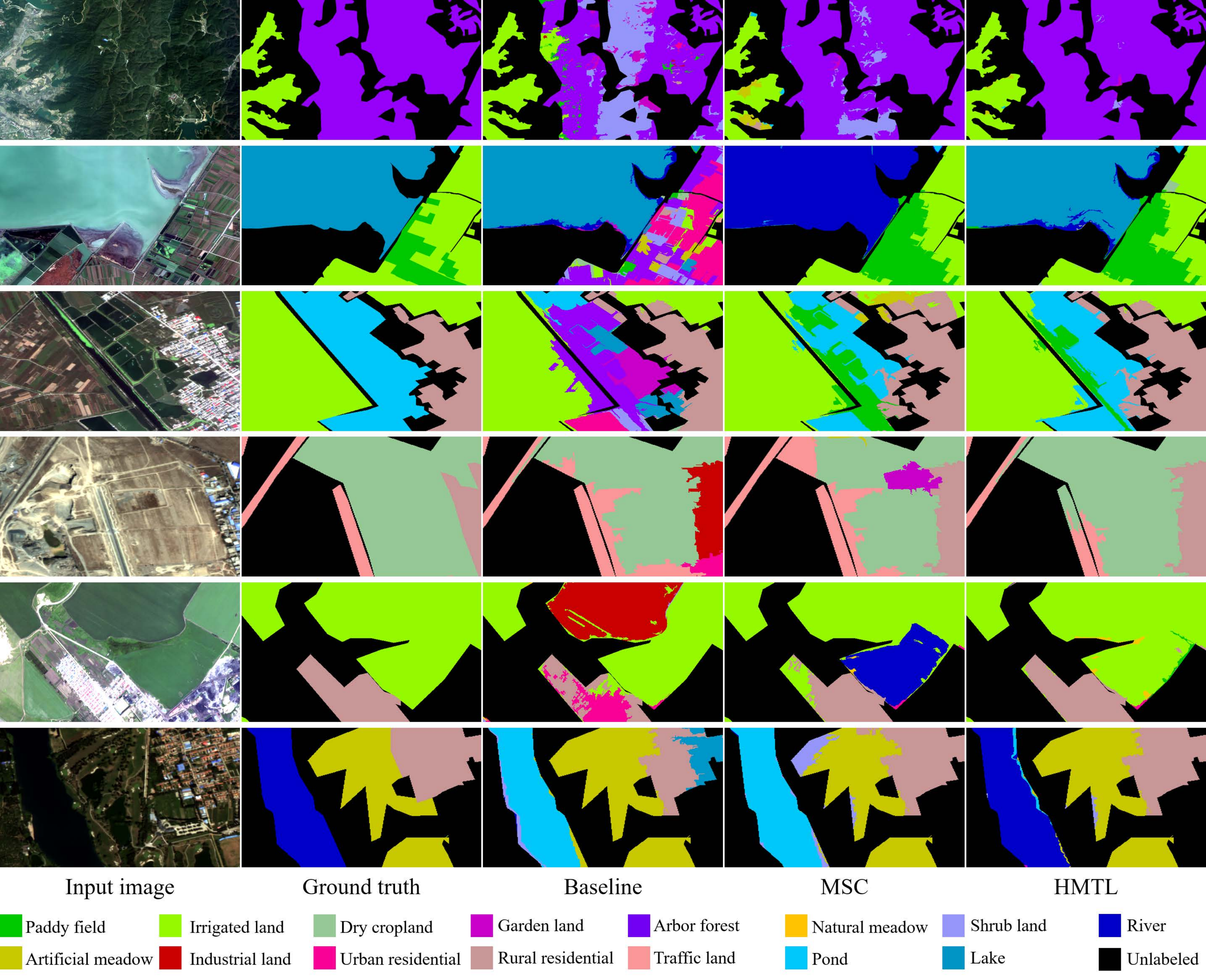}
    \vspace{-4mm}
    \caption{Qualitative comparisons among different classification schemes. The first to fifth columns indicate the original images, ground truth annotations, classification maps from baseline, MSC, and the full implementation of our method, respectively.}
    \label{fig:QCModule}
    \vspace{-4mm}
\end{figure*}

  \subsubsection{Experimental results}
   \textbf{Ablation study:} For multi-task learning, the weights of different tasks can make a big influence on the classification performance. Table~\ref{tab:AblationWeight} shows the classification result under different weight setups, where the MSC strategy is embedded in the baseline network for tile-level scene classification on Million-AID ($T_m$) and pixel-wise semantic parsing on GID ($T_g$). Figure~\ref{fig:MTLossAblation} illustrates the corresponding changes of training losses with respect to different setups of learning weights. As can be seen, the performance of $T_g$ increases gradually as $\mu_g:\mu_m$ changes from $0.1:0.9$ to $0.5:0.5$. This makes sense because the model tends to optimize $T_m$ when $\mu_g$ is smaller than $\mu_m$. As $\mu_g$ increases, the model pays incremental attentions to optimize the $T_g$ and borrows knowledge learned from Million-AID concurrently. Particularly, when $\mu_g:\mu_m=0.1:0.9$, the model can be well optimized for $T_m$, and thus, $T_m$ achieves the best performance with this setup (Figure~\ref{fig:MTLossAblation}(a)). When $\mu_g:\mu_m=0.3:0.7$, the optimization for $T_m$ is slightly insufficient while $T_g$ gains significant improvement as shown in Figure~\ref{fig:MTLossAblation}(b). When $\mu_g:\mu_m$ changes to $0.5:0.5$, both $T_g$  and $T_M$ can be well optimized (Figure~\ref{fig:MTLossAblation}(c)) and $T_m$ can also benefit from the knowledge learned from GID. Thus, the model gains obvious performance improvement. However, as $\mu_g:\mu_m$ continues to increase, the performance of both $T_g$ and $T_m$ begin to decline because there is a risk of overfitting for $T_g$ and insufficient optimization for $T_m$ as shown in Figure~\ref{fig:MTLossAblation}(d). Generally, the best performance for $T_g$ can be acquired when $\mu_g=\mu_m=0.5$, which is employed in subsequent experiments. As our purpose is to explore the possibility of transferring knowledge in Million-AID to GID for semantic classification of land cover, we will focus on reporting the performance of the main task (\emph{i.e.}, pixel-wise semantic classification on GID) in the following context.
 
   For a better understanding of our presented hierarchical multi-task learning method, detailed ablation studies are conducted with different module settings. Specifically, the ResNet50 is employed as the baseline as introduced before. Then we gradually attach the multi-task scene classification (MSC), hierarchical attention network (HAN), and hierarchical semantic integration (HSF) to the baseline model. Table~\ref{tab:AblationModule} shows the performance comparison of different setups tested on GID. As can be seen, the results achieved by the baseline is far from satisfactory due to the sparsity of training samples. When employing the MSC strategy, the classification performance reaches 72.38\% of OA, 42.71\% of mIoU, and 66.65\% of Kappa, which are 13.29\%, 11.92\%, and 15.06\% higher than those of baseline, respectively. This strongly verifies the effectiveness of MSC, where diverse scene samples of Million-AID can bring implicit data augmentation and greatly boost the semantic feature learning for GID content representation. Under the circumstances, semantic knowledge contained in Million-AID can be effectively transferred to improve the performance of land cover classification on GID.  
   
   When the HAN is introduced, the classification is conducted with the hierarchical features from different streams, respectively. For a scene image, the highest classification score among different streams is adopted to output the corresponding semantic category. As shown in Table~\ref{tab:AblationModule}, the classification performance is further improved with HAN. This mainly benefits from the hierarchical attention mechanism, where the essential features of a specific scene can be learned within individual layers. With the HSF scheme integrated, the advantage of attention features at different levels is significantly improved and the performance reaches 73.03\% of OA, 43.68\% of mIoU, and 67.33\% of Kappa. As can be seen, the MSC strategy helps most in improving the classification performance while the full implementation of our method achieves the best result. 
    
    \begin{table}
        \centering
        \caption{Performance comparison of different setups}
        \setlength{\tabcolsep}{1.7mm}{
        \begin{tabular}{cccc|ccc}
            \hline
            Baseline &MSC  &HAN  &HSF  &Kappa (\%)  &OA (\%)   &mIoU (\%)         \\
            \hline
            \hline
            % \checkmark  &           &           &        &49.29   &61.13       \\
            \checkmark  &           &           &      &51.59     &59.09   &30.79  \\
            % \checkmark  &\checkmark &           &        &63.44   &72.76       \\
            \checkmark  &\checkmark &           &      &66.65    &72.38   &42.71  \\
            % \checkmark  &\checkmark &\checkmark &        &63.45   &72.97       \\
            \checkmark  &\checkmark &\checkmark &      &66.79     &72.52  &43.07  \\
            % \checkmark  &\checkmark &\checkmark &\checkmark &\textbf{64.04} &\textbf{73.50}     \\
            \checkmark  &\checkmark &\checkmark &\checkmark  &\textbf{67.33}  &\textbf{73.03}    &\textbf{43.68}   \\
            \hline
        \end{tabular}}
        \label{tab:AblationModule} 
        \vspace{-3mm}
    \end{table}

   Figure~\ref{fig:QCModule} shows the qualitative comparisons of different classification schemes. It is shown that similar categories are easy to be confused by the baseline method. By employing the MSC strategy, many confusing categories can be distinguished, which verifies the effectiveness of the multi-task learning for transferring the scene knowledge contained in Million-AID. With the full implementation of our developed method, the misclassification within some local areas is further corrected, which is consistent with the performance improvement in Table~\ref{tab:AblationModule}. In general, the designed modules greatly help to grasp the essential semantic knowledge of scene images in Million-AID and GID, thus, improve the generalization ability of the semantic classification model. 
%   All of the above-designed mechanisms help to grasp the essential semantic knowledge of scene images in Million-AID and GID, thus, improve the generalization ability of the semantic classification model. 
        
  \textbf{Performance comparison:} The presented method is compared with several object-based classification methods provided by~\cite{Tong2019GID}, where four typical features including spectral feature (SF), co-occurrence matrix (GLCM), different morphological profiles (DMP), and local binary patterns (LBP) were fused to obtain the scene representation denoted as SGDL for simplicity. Then, maximum likelihood classification (MLC), random forest (RF), support vector machine (SVM), and multi-layer perception (MLP) are used as classifiers for scene classification, respectively. Besides, we compare our method with the CNN model pretrained on the~{\em large-scale classification set} of GID (PT-GID)~\cite{Tong2019GID}. For comprehensive comparison, the presented method was also compared with the end-to-end semantic segmentation models, such as U-Net~\cite{UNet2015}, PSPNet~\cite{PSPNet2017}, DeepLab V3+~\cite{chen2018encoder}, and its variations like DeepLab V3+ Mixed loss function (DeopLab V3+ MLF), DeepLab v3+ MobileNet provided by~\cite{ren2020full}. 
  
\begin{table}
    \centering
    \caption{Comparison of classification accuracy among different methods on GID}
    \small
    \setlength{\tabcolsep}{5.3mm}
    \begin{tabular}{l|c|c}
    \hline
    Methods   &Kappa  &OA (\%) \\  
    \hline
    \hline
    MLC + SGDL               &0.145  &23.61    \\
    SVM + SGDL               &0.148  &23.92    \\
    MLP + SGDL               &0.199  &30.57    \\
    RF + SGDL                &0.237  &33.70    \\
    DeepLab V3+ Mobilenet    &0.357  &54.64    \\
    U-Net                    &0.439  &56.59    \\
    PSPNet                   &0.458  &60.73    \\
    DeepLab V3+              &0.478  &62.19    \\
    DeepLab V3+ MLF          &0.598  &69.16    \\
    PT-GID~                  &0.605  &70.04    \\
    \textbf{Ours}            &\textbf{0.673}  &\textbf{73.03}   \\
    \hline
    \end{tabular}
    \label{tab:TransferGID} 
    \vspace{-3mm}
\end{table}

  The quantitative results of different methods are summarized in Table~\ref{tab:TransferGID}. As can be seen, our method significantly outperforms the object-based ones, showing the superiority of our presented method for semantic content understanding of aerial images. The best result achieved by image segmentation models is 59.8\% of Kappa and 69.16\% of OA from DeepLab V3+ Mixed Loss Functions (MLF). Note that the segmentation models and its variations are based on fully convolutional networks which learn pixel-wise semantics in an end-to-end way. Nevertheless, our method achieves 7.5\% higher Kappa and 3.87\% higher OA than those achieved by DeepLab V3+ MLF, indicating the effectiveness of our method for pixel-wise semantic parsing of aerial images. Particularly, PT-GID was achieved by transferring knowledge in the~\emph{large-scale classification set} of GID, which contains 150k samples relevant to the~\emph{fine classification set} of GID. Thus, PT-GID achieves remarkable performance on the fine land-cover classification set. Despite this, our presented method achieves 6.8\% higher Kappa and about 3\% higher OA, showing the strong transferability and effectiveness of our presented method.
    
 \begin{figure*}
    \centering
    \includegraphics[width=1\linewidth]{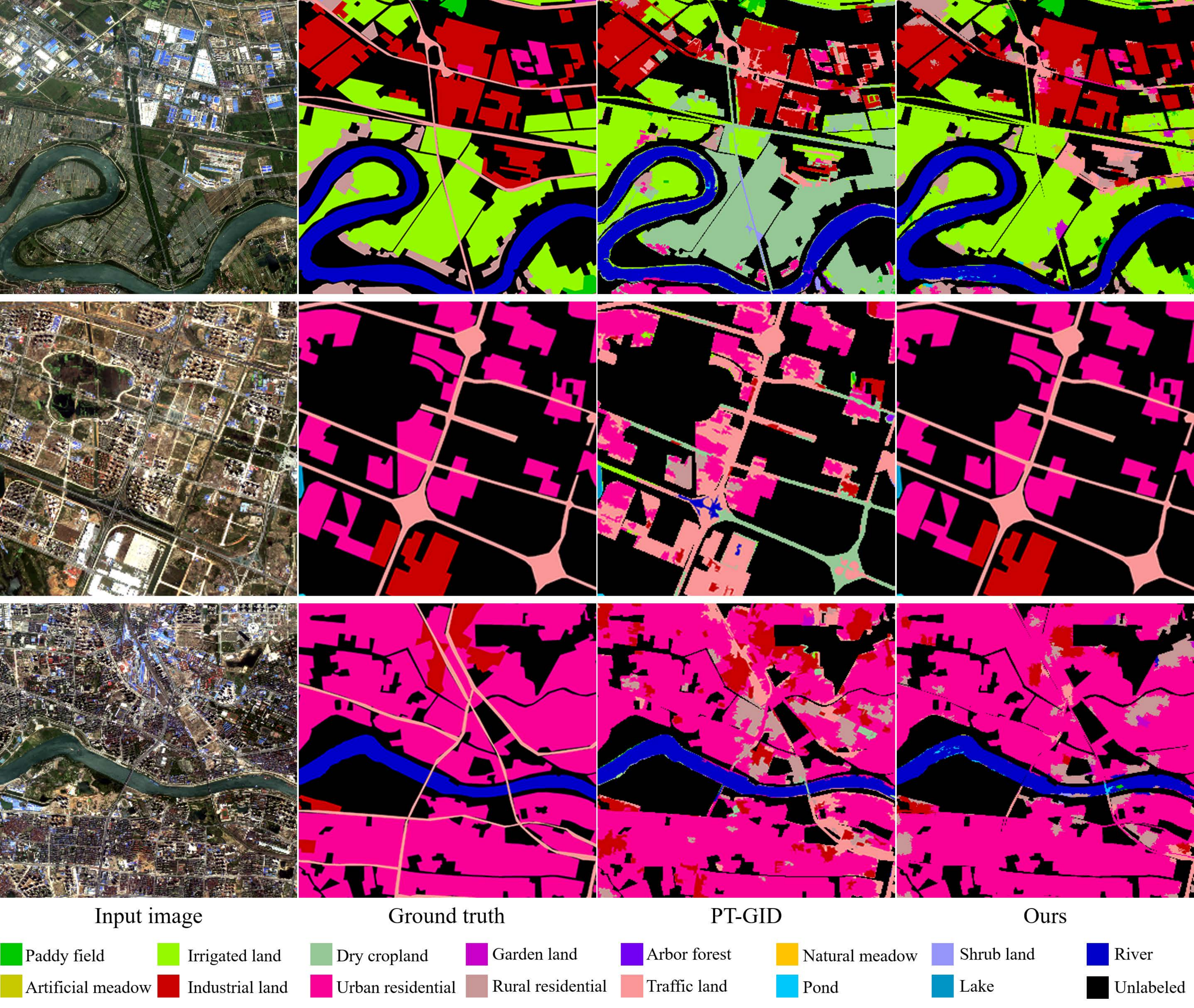}
    \vspace{-4mm}
    \caption{Visualization of the land cover classification results on the~\emph{fine classification set} of GID. Images in the first to fourth columns indicate the original image, ground truth annotations, classification maps of PT-GID~\cite{Tong2019GID}, and classification maps of our method, respectively.}
    \label{fig:CMSOTA}
    \vspace{-4mm}
 \end{figure*}

 Figure~\ref{fig:CMSOTA} provides the intuitive visualization of the pixel-wise classification results on the~\emph{fine classification set} of GID. To save space, we compare the best two results achieved by PT-GID and our designed method. As can be seen from the first row, the~\emph{irrigated land} is heavily misclassified as~\emph{dry cropland} by PT-GID because of the difficulty in distinguishing their similar visual features, such as the texture and structural information. By contrast, our method can discriminate the~\emph{irrigated land} more accurately even it is widely distributed. This benefits from our hierarchically fused features, which can simultaneously incorporate the visual features and semantic information toward specific scene content. In city areas shown in the second row, many semantic categories, such as the~\emph{traffic land},~\emph{urban residential},~\emph{industrial land}, and~\emph{dry cropland}, are heavily confused by PT-GID while our method obtains more accurate classification result.
%  Particularly, information extraction for~\emph{traffic land} is far from desirability. In fact, the proportion of~\emph{traffic land} in city areas is small. Thus, it can be easily confused with adjacent areas and bring the category imbalance problem while learning an interpretation model. 
 This contributes to the semantic attention feature learning in our method, which helps to grasp the essential information for discriminating content of different categories. Likewise, the extraction of~\emph{urban residential} areas is significantly improved by our method as shown in the third row. On the whole, the classification maps of our method present more homogeneous areas and provide more smoother classification result than those of PT-GID. Thus, our method provide much better classification result than the others, which is consistent with the result in Table~\ref{tab:TransferGID}.  

\section{Conclusions} \label{Conclusions}
In this paper we address aerial scene parsing from tile-level scene classification to pixel-level semantic Labeling. Specifically, a review of aerial image interpretation was firstly conducted from its development perspective. It is shown that the interpretation prototype of aerial images has been progressing with the improvement of image resolution and experienced the stages from pixel-wise image classification, segmentation-based image analysis, and tile-level image understanding. Then, we detailed the large-scale dataset, i.e., Million-AID, to be released for aerial scene recognition. Intensive experiments with popular CNN frameworks indicate that Million-AID is a challenging dataset which can be employed as a benchmark for multi-class and multi-label aerial scene classification. Fine-tuning CNN models pretrained on Million-AID show consistent superiority than those pretrained on ImageNet for aerial scene classification, which demonstrates the strong generalization ability of Million-AID. Besides, we designed a hierarchical multi-task learning framework and achieved the state-of-the-art result for pixel-wise semantic labeling, which is an profitable attempt to bridge the tile-level scene classification toward pixel-wise semantic parsing for aerial image interpretation. 

In the future work, we will dedicate our efforts to enrich the Million-AID with more semantic categories and expand the scale of aerial scene images. Knowledge transfer by Million-AID will also be extended to other related tasks, such as object detection and semantic segmentation, to further explore the transferability of Million-AID. We hope that is work can enhance the development of content interpretation algorithms in the field of remote sensing.  

{\small
\bibliographystyle{IEEEtran}
%\balance
\bibliography{refs.bib}
}

\end{document}